\documentclass[11pt,a4paper]{article}

\usepackage[T1]{fontenc}
\usepackage[utf8]{inputenc}
\usepackage{lmodern}
\usepackage[top=2.5cm,bottom=2.5cm,left=2.5cm,right=2.5cm]{geometry}
\usepackage{amsmath}
\usepackage{amssymb}
\usepackage{graphicx}
\usepackage{booktabs}
\usepackage{array}
\usepackage{multirow}
\usepackage{enumitem}
\usepackage{caption}
\usepackage{hyperref}
\usepackage{xcolor}
\usepackage{microtype}
\usepackage{parskip}
\usepackage{titlesec}
\usepackage{longtable}
\usepackage[authoryear,round]{natbib}

\hypersetup{colorlinks=true,linkcolor=blue,citecolor=blue,urlcolor=blue}

\titleformat{\section}{\large\bfseries}{\thesection.}{0.6em}{}
\titleformat{\subsection}{\normalsize\bfseries}{\thesubsection.}{0.6em}{}
\titleformat{\subsubsection}{\normalsize\bfseries\itshape}{\thesubsubsection.}{0.6em}{}

\titleformat{\paragraph}[block]{\normalsize\bfseries\itshape}{}{0em}{}
\titlespacing*{\paragraph}{0pt}{1.2ex plus 0.5ex}{0.5ex}

\begin{document}

\graphicspath{{Figures/}}

\begin{center}
  {\Large\bfseries LCAi: Life Cycle Assessment with big data fusion and
    retrieval-augmented generation-assisted interpretation\par}
  \vspace{1.2em}
  Georgios Tsironis$^{1,2}$, Juan D.\ Medrano-Garc\'{i}a$^{1,2}$,
  Gonzalo Guill\'{e}n-Gos\'{a}lbez$^{*,1,2}$
  \vspace{0.6em}

  {\small
    $^1$Institute for Chemical and Bioengineering, Department of Chemistry and Applied Biosciences,\\
    ETH Z\"{u}rich, Vladimir Prelog Weg~1, Z\"{u}rich 8093, Switzerland\\
    $^2$NCCR Catalysis, Zurich 8093, Switzerland\\[0.3em]
    $^*$Corresponding author.\\
    \texttt{gonzalo.guillen.gosalbez@chem.ethz.ch}
  }
\end{center}
\vspace{1em}

\noindent\textbf{Abstract}

\noindent\textbf{Purpose}

The interpretation phase of life cycle assessment (LCA) often lacks structured mechanisms for translating
quantified improvement opportunities addressing environmental hotspots into actionable strategic pathways under
technological, social, and policy uncertainty. To overcome this limitation, this study introduces a
perspective-conditioned retrieval-augmented generation (RAG) framework for LCA interpretation, where a
multi-perspective retrieval and controlled synthesis is incorporated in the artificial intelligence
(AI)-assisted LCA.

\noindent\textbf{Methods}

To operationalise large language models (LLMs) in LCA interpretation, a perspective fusion RAG architecture
was developed, covering academic, industry, public discourse, and European union (EU) funding datasets. Our
approach comprises three steps: (1) a scenario anchor defining system boundaries and decarbonization targets,
(2) a set of perspective-specific micro-queries with constrained retrieval, and (3) a neutral synthesis step
integrating only ledger-stored outputs without further retrieval. The framework is demonstrated through a
hydrogen-enabled diesel reduction use case in an Italian apple production facility using GPT-5 nano as the
reasoning model.

\noindent\textbf{Results}

Considering the potential points of interest for a future stakeholder, our framework generates an auditable
2030 roadmap without introducing ungrounded claims. More specifically, academic sources illustrate system-level
constraints and efficiency trade-offs, while industry sources highlight deployment pathways and business model
aspects. Public discourse revealed legitimacy and safety concerns, and EU policy data identified funding
accelerators and ecosystem structures. Overall, the structured retrieval and constrained synthesis are designed
to mitigate the risk of hallucination while preserving cross-domain diversity. The approach presented can support more disciplined translation of impact results into strategic pathways and opens up new avenues for
the use of advanced AI tools in LCA studies, particularly those focused on technologies that could be deployed
at scale.

\noindent\textbf{Conclusion and recommendation}

This proof-of-concept demonstrates how AI-assisted, evidence-grounded interpretation can support
implementation-oriented decision-making beyond conventional LCA studies. Future work will focus on analytical
extensions to better inform implementation-support practices. To further evaluate the feasibility of the
proposed roadmaps, future research could integrate more public and private data, including, for instance,
sustainability reports, legislation or ISO standards. Moreover, the framework may also be extended toward
multi-objective LCA scenarios and dynamic policy environments.


\section{Introduction}

The concept of sustainability is heavily related to reducing environmental impacts, which requires assessing
them in products and services over the life cycle precisely \citep{Ross2002}. In this context, life cycle
assessment (LCA) is the prevalent tool for systematic investigation and evaluation of industrial systems
\citep{Lewandowska2011}. Since the early stages of its creation, the aim of LCA was to provide the
appropriate insights to facilitate process selection, design, and optimization \citep{Keoleian1993,Azapagic1999}.
More specifically, the last phase of the LCA~--- the interpretation phase~--- blends the technical analysis with the
real world in terms of choices and actions needed toward achieving the suggested outcomes \citep{ISO14044}. In this context, artificial intelligence (AI) is expected to help bridge the gap between academic research and industrial practice \citep{Kumar2026}.
For example, decarbonizing the energy sector is further affected by existing social mechanisms with people's
support or resistance, playing a crucial role in the change of speed and type \citep{Verrier2022};
consequently, these mechanisms should be considered when addressing hotspots in the LCAs of such systems.

With the advent of generative AI, we have recently witnessed an increasing number of studies exploring LCA
automation using AI methods in the various LCA phases \citep{Preuss2024}. AI applications in this field include the development
of AI-driven software for automated life cycle inventory construction for buildings \citep{Petrosa2025}, and
the use of ensemble AI models for predicting sustainability performance in vehicles parts production
\citep{Shafiq2024}. Additionally, diverse AI methods were also applied for predicting the footprint of
agriculture products in soybean cultivation \citep{Mohammadi2023}. AI-based automation has been reported as
very promising in terms of time and resource efficiency, despite lacking scientific quality assessment and
norms for use \citep{Preuss2026}. Interestingly, the literature indicates the dominance of supervised machine
learning (ML) algorithms mainly for data collection and inventory analysis. Notwithstanding these efforts,
AI-in-LCA technologies often lack real-world validation and standardization in sectors like building
retrofitting \citep{Luan2026}.

More recently, the rapidly developing AI landscape offers further opportunities for automating LCA based on
large language models (LLMs) and generative algorithms \citep{Nwagwu2025}, which we further explore in this
work. These LLMs are now underpinned by algorithms for optimizing performance and reducing hallucination,
i.e., irrational and uncontrolled behaviours of LLMs \citep{Rawte2023}, which in the context of LCA could
lead to wrong advice. These LLMs improvements include methods for optimising training data \citep{Lin2024},
and to support the fine-tuning of the LLMs, for example by re-training them based on proprietary data
\citep{J2024}. For example, previous studies have investigated medical fine-tuned LLMs, showing improved
performance in question-answer (QA)-based prompts \citep{Ji2023}. Hallucination causes can be extrinsic,
i.e., output not aligned with the training data, and intrinsic, i.e., model's failure to understand the
prompt given \citep{Sekrst2025}. Regardless of its nature, hallucination can hamper the successful use of
LLMs in science and engineering, and recent works investigated extrinsic hallucinations and refusal rates of
different models \citep{Bang2025}.

The retrieval-augmented generation (RAG) framework was recently introduced to tackle hallucinations and
enhance language generation tasks \citep{Lewis2021}. Relying on vectorised text chunks and prompt-matching
retrieval, RAG can provide very specific insights or information regarding the user prompt. Additionally, it
offers LLM modularity and database scalability in a dynamic way. RAG was applied across different domains and
sectors, mostly on businesses relying on information extraction from their private data for operational
optimisation \citep{Arslan2024}. Previous applications focused on the cybersecurity and governance domains
with knowledge transfer \citep{Hasan2026}, the healthcare sector \citep{Neha2025,AboElEnen2025}, and in the
educational domain \citep{Swacha2025}. In the context of LCA, LLMs and RAG were applied to LCA
optimization. For example, RAG was also used for retrieving and analysing LCA users' QAs that can effectively
provide relevant context for the reporting stage \citep{ZhangX2025}. Further, ChatGPT was also applied for
generating more understandable design knowledge from LCA reports with different categories of prompts
\citep{Goridkov2024}. 

Our work differentiates itself by introducing a novel approach to perspective fusion
using big data from diverse opinion-based, networking, and scientific platforms. In this context, RAG does not
serve as a seeker of absolute ground-truth, but rather as a selective retriever of the most relevant
information within a vast data pool. Ultimately, anchoring this process with LCA data inputs, followed by a
multi-perspective synthesis, provides an innovative workflow with significant scalability potential. Following this approach, here we  apply RAG in the interpretation phase of LCA, which quantifies the impact of a set of
alternatives, often called scenarios, that are evaluated in terms of environmental performance using metrics. In the interpretation phase, such alternatives are analysed thoroughly, identifying
hotspots that could be tackled via targeted actions and discussing the potential occurrence of burden-shifting.
However, the LCA interpretation phase usually lacks in-depth feasibility and step-by-step implementation
roadmaps that could enable knowledge transfer to management \citep{Prado2022}. Hence, it could be significantly enhanced by broadening the scope of the analysis and combining technical and
non-technical information. Here, we accomplish this goal by systematically
integrating heterogeneous and socio-technical evidence under controlled traceable AI reasoning. More
precisely, we present a methodological approach that leverages LLMs and online data platforms for providing
structured implementation-oriented pathways downstream of conventional LCA interpretation. Our work explores
whether AI-assisted evidence synthesis can function as an analytical extension that strengthens
implementation-oriented decision support. Sufficiently matured and validated, such approaches open new
avenues for future methodological contributions in LCA practice.

From a methodological viewpoint, a pilot graphical user interface (GUI) was developed where the user can
directly interface with the analytical pipeline, including the GPT-5 nano via the OpenAI API. The
exploitation of diverse data from online platforms, combined with the overall synthesis of the prompt outputs
assists users in the interpretation of the LCA outputs and provides tangible implementation pathways to the
targeted stakeholders, including public and private sector stakeholders needing insights on the strategic
management, decision-making, and business-model design. We illustrate the capabilities of the approach in a
proof-of-concept case on ``green hydrogen'', ``renewable hydrogen'', and ``renewable feedstocks''. Our
results demonstrate the value of evidence-based reasoning with LLM agents to support more comprehensive and
insightful LCAs. Overall, the multi-perspective data alongside the strict output instructions enable a more
tangible roadmap and pathway definition for implementing systemic changes toward reducing environmental impact.

\section{Methodology}

\begin{figure}[h]
  \centering
  \includegraphics[width=0.82\textwidth]{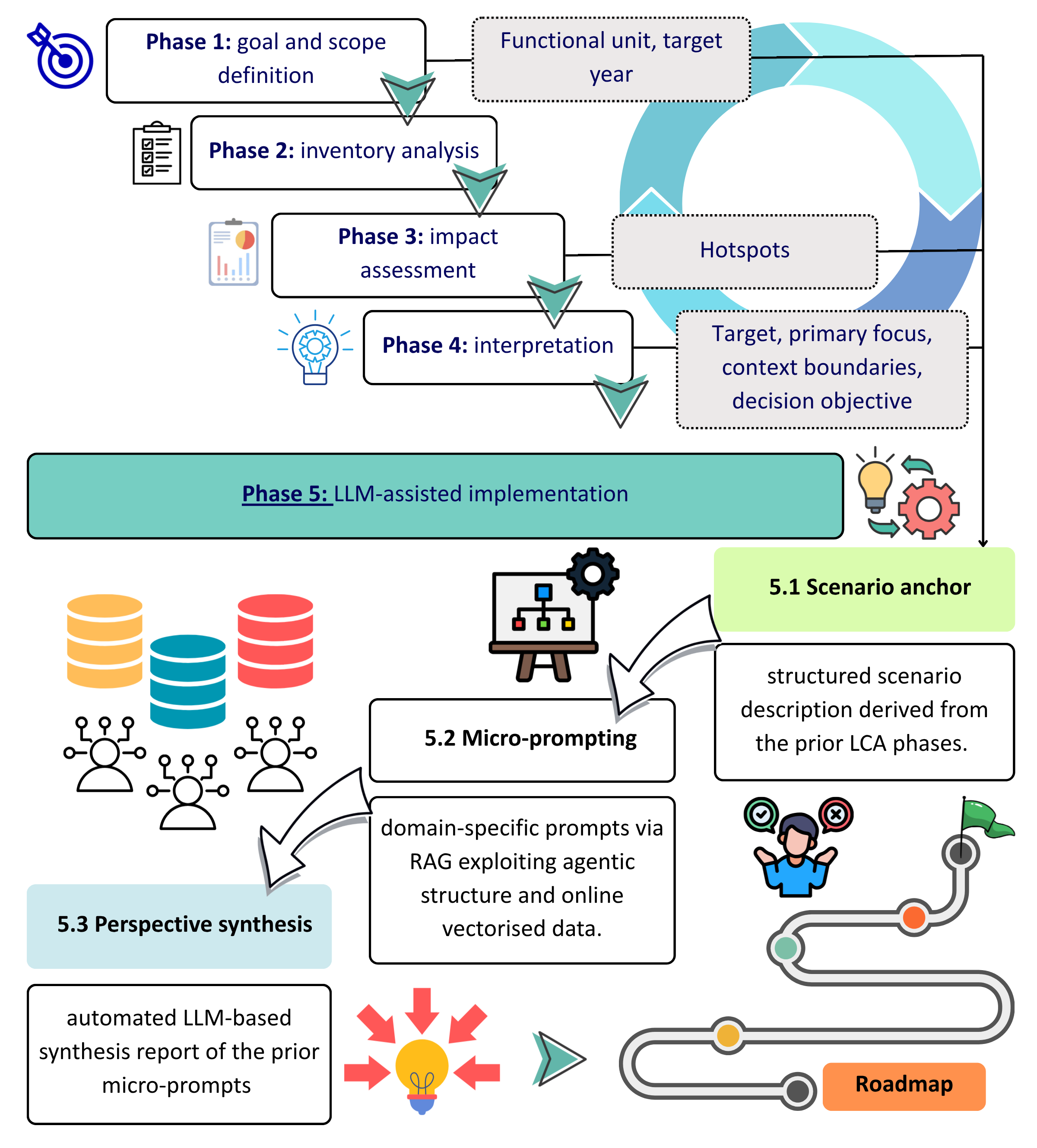}
  \caption{Methodological framework depicting an implementation-oriented analytical extension downstream of
    the conventional four LCA phases.}
  \label{fig:1}
\end{figure}

In essence, we follow the same LCA phases described in the ISO~14040/44 \citep{ISO14044}, where in phase
four we integrate an AI-assisted framework to improve the results interpretation based on a corpus of
relevant data. More precisely, the proposed methodological framework is developed as an extension to the four
LCA phases, where the goal is to further contextualise the LCA results and study implementation pathways
based on the scenarios analysed (\textbf{Figure~\ref{fig:1}}). In this fashion, the information flow
initiates from the first LCA phase~--- goal and scope definition~--- which provides the functional unit,
system boundaries and the target year of the analysis. After the inventory and impact assessment phases are
completed, the main hotspots are identified in phase four, providing essential data for applying our RAG
approach. This fourth phase also provides the target audience, primary focus, context boundaries, and
decision objective(s), which define the anchoring context for the proposed AI workflow. This anchoring
enables the appropriate connection between the LLM and our technical LCA work.

The methodological framework employs RAG for the LCIA results contextualization through data injection from
online platforms providing different perspectives reflecting a range of stakeholders relevant to the case
study. In the next step, we detail the data collection and preprocessing. Subsequently, the detailed
three-step LLM-assisted pipeline is presented.

\subsection{Data collection \& preprocessing}

As a basic requirement for a perspective-conditioned retrieval, four heterogeneous corpora are constructed
representing distinct knowledge domains relevant to sustainability decision-making: academic literature
(Scopus abstracts), industry discourse (LinkedIn company profile descriptions), EU funding \& innovation
policy (CORDIS project objectives), and public discourse (YouTube comments, Reddit threads and sub-threads,
and Bluesky posts) with adjusted queries per platform. These sources are selected to capture multiple
information lenses including among others, peer-reviewed technical feasibility, market deployment signals,
institutional acceleration mechanisms, and societal perception narratives. For the use case developed in
this work, as a starting point, we focus on the hydrogen sector and, more specifically, green and renewable
hydrogen, and renewable feedstocks. Data collection employs structured queries (Table~\ref{tab:1}), and
data are exported in tabular formats (csv/xlsx) for standardised processing. The datasets can be scaled in
numbers and across platforms or diverse domains and concepts.

\begin{table}[h]
  \centering
  \caption{Dataset description per platform and perspective.}
  \label{tab:1}
  \small
  \begin{tabular}{llp{4.2cm}rp{2.6cm}}
    \toprule
    \textbf{Perspective} & \textbf{Platform} & \textbf{Search term} &
    \textbf{Documents} & \textbf{Data collection method} \\
    \midrule
    \multirow{3}{*}{Academic} & \multirow{3}{*}{Scopus}
      & ``green hydrogen''       & 7,460  & \multirow{3}{*}{Elsevier API} \\
      & & ``renewable hydrogen''  & 1,517  & \\
      & & ``renewable feedstocks'' & 2,111 & \\
    \midrule
    Business & LinkedIn & ``hydrogen'' & 4,834 & Web scraping tool \\
    \midrule
    \multirow{6}{*}{Public}
      & YouTube & ``green hydrogen'' AND ``renewable hydrogen'' & 30,318
      & Google APIs / youtube\_v3 \\
      \cmidrule{2-5}
      & \multirow{3}{*}{Reddit}
      & ``green hydrogen''        & 18,929 & \multirow{3}{*}{Reddit API} \\
      & & ``renewable hydrogen''  & 23,705 & \\
      & & ``renewable feedstocks'' & 311   & \\
      \cmidrule{2-5}
      & \multirow{3}{*}{Bluesky}
      & ``green hydrogen''        & 32,926 & \multirow{3}{*}{Bluesky API} \\
      & & ``renewable hydrogen''  & 6,975  & \\
      & & ``renewable feedstocks'' & 562   & \\
    \midrule
    EU innovation/funding & Cordis EU & ``hydrogen'' & 2,183 & Order \& download \\
    \bottomrule
  \end{tabular}
\end{table}

The search queries were tailored to the different platforms. For example, the LinkedIn search returned only
110 company profiles for the ``green hydrogen'' query, worldwide. Accordingly, we explored ``hydrogen'' as a
wider search term in company profiles dividing the process into individual searches with geographical
criteria. More specifically, the searches include the major countries in terms of numbers and geographical
coverage while also considering other sectors in the hydrogen LinkedIn landscape (Table~S1).

Preprocessing ensures semantic consistency and retrieval integrity across corpora. Text cleaning includes
removal of duplicates, null entries, formatting artifacts, and non-informative metadata fields. Each document
corresponds to a text unit, defined as a semantically coherent, self-contained passage such as a single
abstract, company description, project objective, or user post/comment. Metadata (e.g., publication year,
source type, project acronym, organisational descriptors) are also collected and linked to each unit to
support traceability, diversity analysis, and diagnostic evaluation during retrieval. The detailed descriptive
data analytics are enclosed in the SI (Section~1). Importantly, the data collection strictly involves
publicly available data that do not expose any private information during the analysis and results
presentation stage of this work. We omit direct mentions to the root text units used for the model
reasoning. Apart from the CORDIS platform, all the others do provide data visibility without having any
special account permission. For the EU funded project data, we only refer to some project names with respect
to their area of focus and innovation without any further mention of additional information. This approach
allows for the use of OpenAI's API without exposing large data volumes.

\subsection{Three-step workflow architecture}

\begin{figure}[h]
  \centering
  \includegraphics[width=\textwidth]{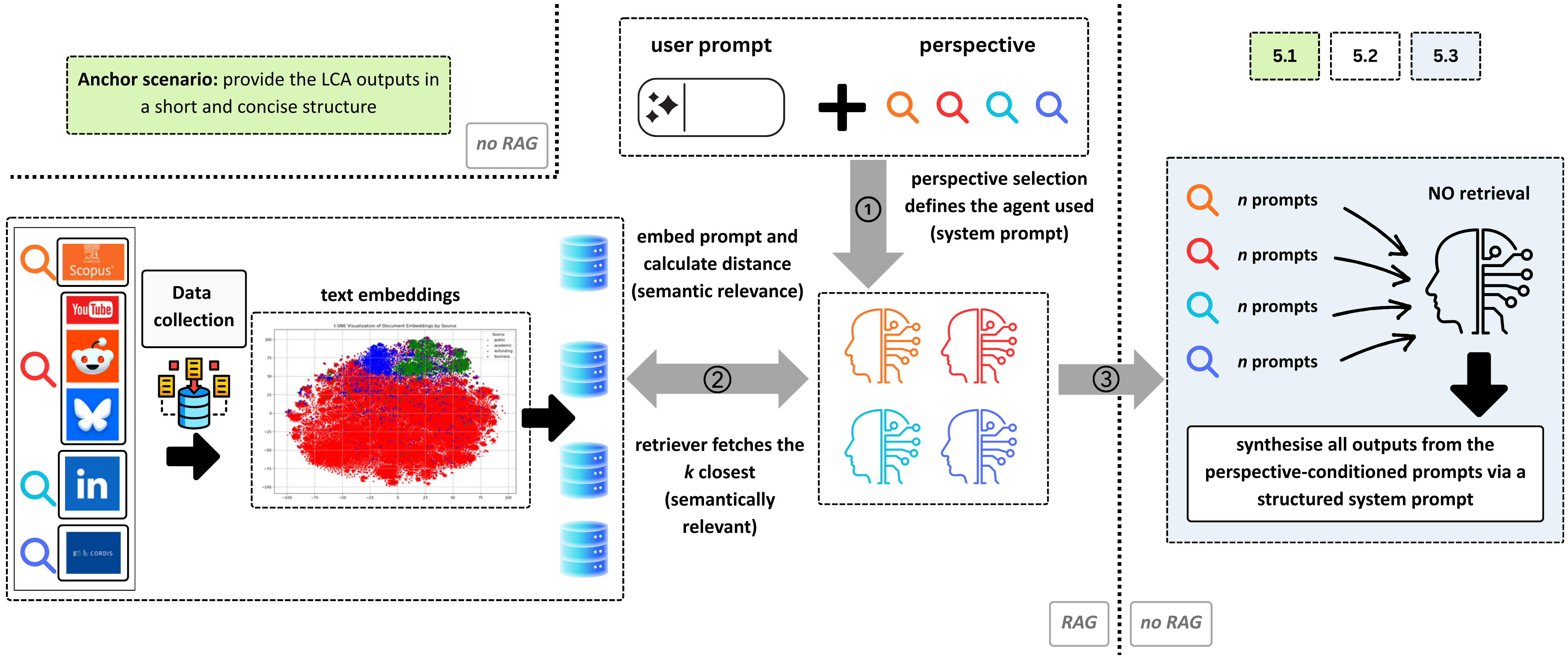}
  \caption{Perspective-conditioned RAG workflow representing the proposed implementation-oriented analytical
    extension to conventional LCA interpretation. The pipeline starts with a structured scenario anchor
    (5.1) derived from the LCA outputs. In step~5.2, user micro-prompts are combined with a selected
    stakeholder perspective (academic, business, public, or EU funding), which determines the corresponding
    vector database and system prompt. The retriever identifies the $k$ most semantically relevant text
    units for evidence-grounded reasoning. Finally, step~5.3 synthesises all perspective-conditioned outputs
    into a consolidated implementation roadmap without additional retrieval.}
  \label{fig:2}
\end{figure}

The proposed framework operates through a structured three-step AI workflow designed to sequential
analytical steps (\textbf{Figure~\ref{fig:2}}). This staged architecture prevents cross-perspective leakage
and enhances interpretative transparency. We create a web-based localhost page using streamlit, a python
package highly suitable for data science projects and easy visualisation of workflows in real time
\citep{Khorasani2025}. This allows us to easily experiment with multiple prompts, effectively change
perspectives, and after finishing the prompting session simply synthesise everything in a single LLM output
using a structured report.

\subsubsection{Contextual anchoring (5.1)}

The first step of the workflow establishes the scenario anchor that connects the conventional LCA phases
with the AI-assisted pipeline. This step transfers the core outputs of the assessment into a structured
context that constrains the subsequent retrieval and reasoning process. More precisely, the anchor includes
the functional unit, target year, environmental hotspots, transition target, context boundaries, and
decision objective, extracted from the previous LCA phases. Apart from the standard LCA terms, we also
refer to some additional ones for describing the required information derived from the four phases alongside
any additional inputs for the next steps. For example, the transition target refers to the goal set by the
decision-maker based on the technical analysis (e.g., 50\% diesel reduction). Further, the decision
objectives describe the desired characteristics of a meaningful and actionable outcome from the AI-assisted
implementation pathway. Additionally, the anchor already encapsulates the decision scenario under
examination considering that the LCA practitioner has identified a suitable systemic change and wants to
dive deeper into the pathways and barriers towards the implementation.

The purpose of the anchor is to ensure that all generated outputs remain aligned with the original LCA
system and implementation objective. Instead of allowing unconstrained prompting, the framework defines a
fixed scenario that guides the retrieval and interpretation steps. Consequently, the LLM operates as a
context-bounded interpretation assistant rather than a general-purpose conversational model. This approach
improves methodological consistency and reduces the risk of irrelevant or ungrounded outputs. The same
anchor remains fixed across all perspectives, allowing controlled comparison between academic, business,
public, and EU-funding initiatives evidence during the subsequent retrieval and synthesis steps.

\subsubsection{Perspective-conditioned retrieval (5.2)}

\paragraph{Embedding \& indexing architecture}

Aiming for a semantic retrieval across heterogeneous corpora, the collected text units are embedded using
the all-MiniLM-L6-v2 sentence transformers model, selected for its balance between computational efficiency
and semantic representation quality \citep{ZhangJ2025}. This model generates dense vector representations
that capture contextual similarity beyond keyword matching, allowing retrieval of conceptually related
passages across technical, industrial, funding, and public discourse domains. Embeddings are generated at
the atomic text unit level to preserve semantic coherence. In our case, chunking was not necessary due to
the data format, type, and length.

\begin{figure}[h]
  \centering
  \includegraphics[width=0.90\textwidth]{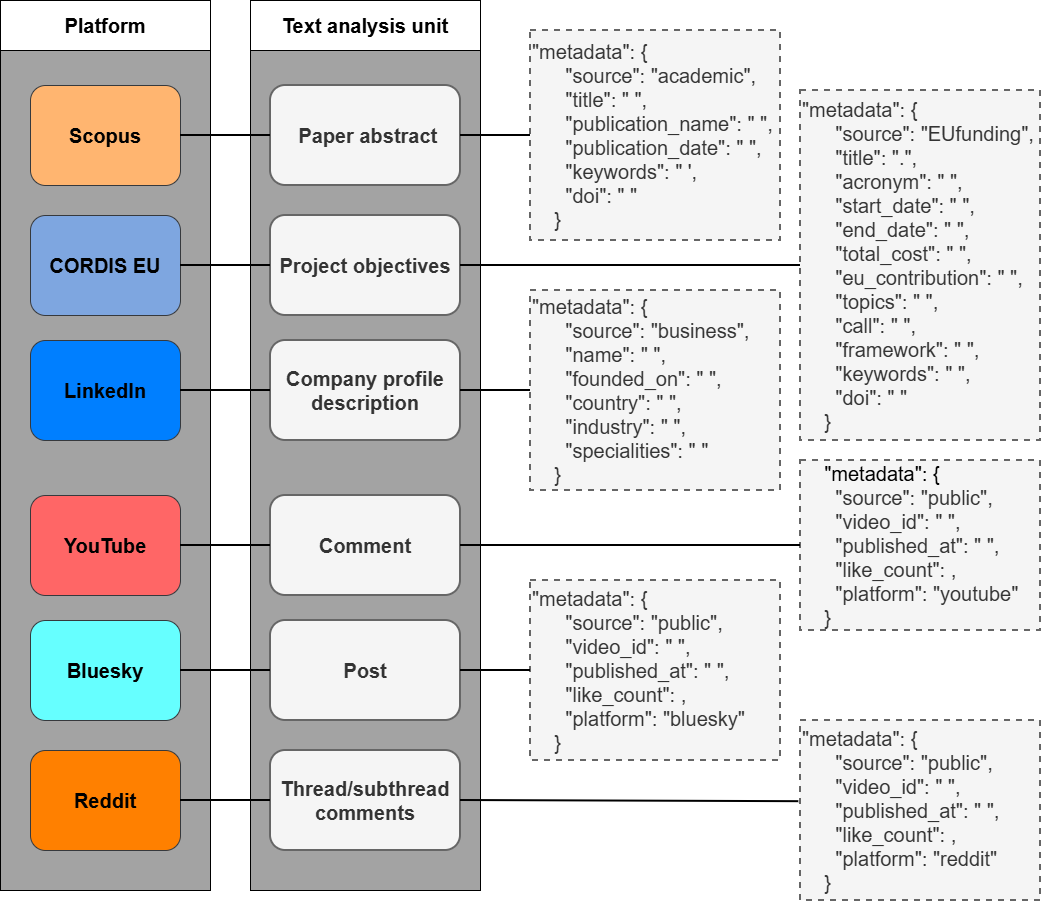}
  \caption{Data models of the individual data source platforms including the core text input complemented
    by the available metadata.}
  \label{fig:3}
\end{figure}

Next, the vector representations are indexed using Facebook AI similarity search (FAISS), enabling efficient
nearest-neighbor retrieval under Euclidean distance \citep{Douze2026}. We construct separate FAISS indices
for each corpus, ensuring strict perspective separation. In the following retrieval step, user queries are
embedded and matched only against the selected perspective index, preventing the unintended blending of
datasets. That means we must select which perspective the pipeline has to utilise for answering the given
prompt. When the perspective is selected, the corresponding dataset, retriever, and LLM system prompt are
used. Further, the data models of each platform contain a free text part which serves as the core of the
RAG pipeline. Apart from this, we also store any complementary information as metadata for more advance
reasoning and a citation-based validation of the outputs (\textbf{Figure~\ref{fig:3}}). For facilitating
the context retrieval, we use a uniform data model for all three public perspective platforms. This explains
the ``video\_id'' field in the Bluesky and Reddit data models, each pointing to the specific post across
platforms.

\paragraph{Micro-prompt design}

The system prompts provide the clear directions needed for the LLM agents with perspective-specific
guidelines. A shared core defines the model's role as an LCA interpretation and strategic decision-support
assistant. This core instruction establishes boundaries for evidence use, limits the invention of facts not
present in retrieved context, requires explicit citation of supporting text units, and mandates separation
between grounded statements and general knowledge. The core prompt also specifies that responses must
directly address the micro-query within the constraints of the predefined scenario anchor (5.1). The
academic perspective emphasises conditional feasibility, system constraints, and uncertainty, while the
business perspective prioritises deployment models and market readiness. On the other hand, the public
discourse perspective foregrounds perception, legitimacy, and skepticism. Finally, the EU policy perspective
focuses on funding instruments, pilot projects, and institutional acceleration mechanisms (\textbf{Section~4
SI}).

The framework's reasoning is governed by a dual-layer system prompting architecture that transitions the
LLM from a general-purpose assistant to a specialised LCA interpretation expert and strategic
sustainability decision advisor. This is achieved through the following technical constraints:

\begin{itemize}
  \item \textbf{Persona-driven heuristics:} The model is initialised with perspective-specific personas,
    each based on the corresponding dataset. These personas do not merely change the tone but shift the
    model's priority toward domain-specific evidence, such as prioritizing technology readiness signals and
    lifecycle trade-offs for academic data, or operational practicality and vendor offerings for business
    data.

  \item \textbf{Contextual improvement and hallucination mitigation:} To ensure evidence-based reasoning
    the model is explicitly prohibited from utilizing internal training data to invent specific figures,
    projects, or dates not present in the retrieved context. If evidence is missing, the model must output
    an `INSUFFICIENT EVIDENCE IN RETRIEVED CONTEXT' flag, effectively suppressing the generation of
    ungrounded or unnecessary responses.

  \item \textbf{Structured output and auditability:} The reasoning process is standardised through a rigid
    output schema. Every claim is required to include a confidence rule score based on evidence density and
    a direct citation in the format `EVIDENCE=[source\_id]'. This ensures that the strategic roadmap is not
    a narrative summary but an auditable ledger tied directly to the scenario's functional unit and
    hotspots.

  \item \textbf{Deterministic scoping via scenario anchors:} By binding every micro-query to a persistent
    scenario anchor the system prompts fine-tune the model's reasoning scope. This prevents the roadmap
    from drifting into generic hydrogen trends and ensures that all synthesised insights remain tethered to
    the specific LCA boundaries defined in the initial assessment.
\end{itemize}

The structured outputs generated across all perspective-conditioned micro-prompts~--- including responses,
quality flags, confidence scores, and evidence citations~--- are collectively stored in a session ledger.
This ledger serves as the sole input to the subsequent synthesis step, ensuring that no new retrieved context
is introduced during final integration. Considering the multi-perspective structure of the pipeline alongside
the diversity in the collected datasets, we construct eight prompts, two per perspective. Further, our paired
prompts aim to check the content availability in each dataset and shed light into online information. This
approach, which serves as a proof-of-concept for the RAG-assisted LCA potential, presents significant
scalability based on the provided data. The number of prompts can vary according to the user's needs and
analysis depth. However, it is important to also consider the LLM's context window which will be used for
the subsequent synthesis step where it must fit all the micro-prompt outputs as the final input for
synthesis.

\paragraph{Quality flags}

Retrieval similarity is computed using Euclidean distance (L2) between the embedded query vector ($q$) and
document vectors ($d$) within a 384-dimensional embedding space generated by the same sentence transformer
model used for text embedding (\ref{eq:L2}).

\begin{equation}
  d_{L2}(q,d) = \|q - d\|_2 = \sqrt{\sum_{i=1}^{384}(q_i - d_i)^2}
  \label{eq:L2}
\end{equation}

Because embeddings were not unit-normalised, L2 magnitude reflects raw vector geometry and depends on
corpus distribution, embedding scale, and query specificity. Consequently, absolute L2 values have no
intrinsic interpretative threshold, for example, a value of 0.5 cannot be universally classified as good
or bad. Instead, L2 distances were used comparatively, both within query with best versus worst retrieved
items, and across queries with mean L2 differences indicating potential retrieval mismatch or corpus
sparsity. Importantly, L2 distance reflects semantic proximity in embedding space but does not directly
measure factual relevance, as no ground-truth labels are available. Retrieval quality therefore relied on
internal consistency diagnostics, including mean and variance of L2, evidence use ratio like the proportion
of cited and retrieved items, citation concentration, and retrieval stability across repeated runs.

The layered prompt design ensures that retrieval remains domain-specific while preserving methodological
transparency. Additionally, responses are required to include structured retrieval quality flags calculated
by the LLM agent itself, such as relevance, coverage, and evidence density, complemented by an explicit
confidence level (Table~\ref{tab:2}). These flags serve as diagnostic signals rather than performance
claims, enabling later analysis of grounding strength, perspective divergence, and confidence calibration.
The LLM uses this self-evaluation approach in the system prompts given for each perspective agent. For a
detailed view of the content refer to the supplementary information (SI) document (Section~4). These
indicators are not computed via deterministic thresholds, rather, they rely on the interpretation of the
retrieved context units. Their purpose is to provide structured transparency regarding the perceived
grounding strength and interpretative robustness of each response. These diagnostics should be interpreted
as structured qualitative signals rather than reproducible performance metrics as minor variation may occur
across repeated runs due to model stochasticity and contextual sensitivity. This topic is also discussed
in the limitations section of the SI (Section~6).

\begin{table}[h]
  \centering
  \caption{LLM-generated response quality flag system for basic output quality.}
  \label{tab:2}
  \small
  \begin{tabular}{p{2.2cm} p{3.8cm} p{8.0cm}}
    \toprule
    \textbf{Quality flag} & \textbf{Values} & \textbf{Description} \\
    \midrule
    Relevance
      & High / moderate / low
      & Reflects the model's assessment of how directly the retrieved text units address the micro-query
        within the constraints of the scenario anchor. It is not derived from similarity scores nor from
        keyword overlap, but from semantic alignment between retrieved evidence and the specific decision
        context. We ask the model to evaluate the relevance of the retrieved units to the prompt given. \\[4pt]
    Coverage
      & Multi-view / single-view / unclear
      & Captures within-perspective diversity of arguments or dimensions present in the retrieved evidence.
        A ``multi-view'' designation indicates the presence of distinct angles, constraints, or supporting
        mechanisms within a single domain. For example, if the 10 retrieved text units for a given prompt
        under the academic perspective present multiple technical limitations, then the coverage score is
        high. \\[4pt]
    Evidence density
      & Strong ($\geq$3 relevant items) / limited (2 items) / sparse (0--1 item)
      & Reflects the model's estimate of how many retrieved units substantively support the generated
        claims. While the retriever always returns a fixed number of top-$k$ items, density is not a simple
        function of $k$. Instead, it represents the proportion of context units meaningfully used in the
        answer. \\[4pt]
    Confidence level
      & High only if evidence $=$ strong and directly matches the micro-query $+$ scenario; medium if
        limited evidence or partial match; low if sparse evidence, weak match, or likely missing
        counterpoints.
      & Synthesises these diagnostics, incorporating relevance, coverage, evidence density, and alignment
        with the scenario anchor. A response may exhibit strong evidence density yet retain medium
        confidence if conclusions are conditional or dependent on system-level uncertainties, which is
        desirable in technically constrained domains. This metric is estimated for each individual point
        raised by the LLM in bullet points. Each prompt response has several reasoning claims, each one
        carrying a confidence score based on the available context (Section~3). \\
    \bottomrule
  \end{tabular}
\end{table}

\subsubsection{Neutral synthesis (5.3)}

The final step of the workflow synthesises the outputs generated during the perspective-conditioned
retrieval step into a consolidated implementation-oriented narrative. More precisely, the synthesis agent
receives only the stored outputs, citations, and quality flags from the previous micro-prompts, without
performing any additional retrieval. This ledger-constrained approach prevents the introduction of new
evidence during the final generation step and maintains traceability across the workflow. The purpose of
this step is to integrate the different stakeholder perspectives into a structured roadmap describing
potential implementation pathways, critical constraints, and areas of convergence or tension across the
retrieved evidence. Consequently, the synthesis step transforms multiple isolated prompt responses into a
unified interpretation of the LCA transition scenario while preserving the contextual boundaries
established by the scenario anchor.

The model's reasoning is conditioned to operate as a neutral mediator that must resolve or highlight
cross-perspective conflicts~--- such as the tension between industry deployment optimism and academic
technical skepticism~--- without introducing outside bias. To ensure the output is actionable for LCA
stakeholders, the prompt enforces a hierarchical roadmap structure and mandates the preservation of all
evidence-based tags. Specifically, the synthesis reasoning must consolidate evidence by aggregating the
overlapping claims from different perspectives to identify high-consensus systemic enablers; maintain
traceability through retain the original source IDs and confidence scores in the final summary to ensure
the roadmap remains fully auditable back to the raw data; finally, filtering generalities and explicitly
suppress generic AI-generated advice by prioritizing findings that are directly tethered to the scenario
anchor.

\section{Case Study}

As a case study for this framework, we analyse how to improve environmentally apple production in Italy.
Apple production was selected as a proof-of-concept because the LCA identified a clear diesel-related
decarbonisation hotspot, while hydrogen substitution introduces substantial socio-technical uncertainty
requiring evidence beyond environmental impact assessment alone. This creates a representative scenario for
demonstrating the value of perspective-conditioned AI-assisted interpretation in translating hotspot
reduction targets into actionable transition pathways.

\begin{figure}[h]
  \centering
  \includegraphics[width=0.60\textwidth]{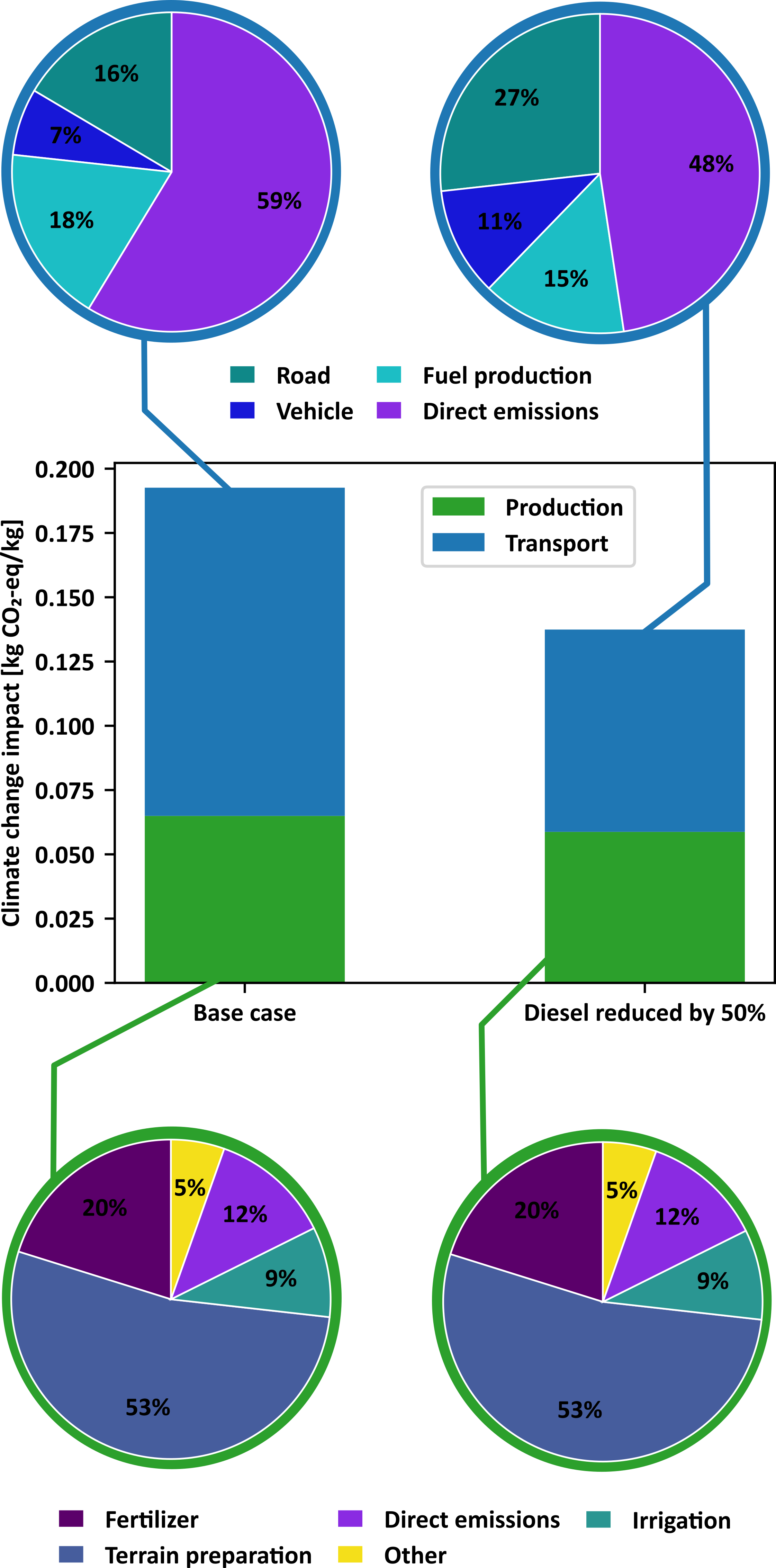}
  \caption{LCIA results of the case study. The ``base case'' scenario consists of the standard production
    of 1~kg of apples in Italy and transport to the main European consumer countries. The ``diesel reduced
    by 50\%'' scenario assumes diesel consumption is halved in order to assess the effect on the climate
    change impact of the system. More details on the case study can be found in the SI Section~2.}
  \label{fig:4}
\end{figure}

The LCA follows the four phases described in the ISO~14040/44 \citep{ISO14044}. For the goal and scope, we
consider a cradle-to-gate assessment with a cutoff attributional approach of the production of apples in
Italy and its subsequent export to the most relevant European country consumers. We use the ecoinvent
database v3.12 to model the background system \citep{Wernet2016}. The chosen functional unit is 1~kg of
apples. The life cycle inventory (LCI) for the foreground system was built using the ecoinvent database
v3.12 for the apple production activity and export data to estimate the average ton$\cdot$km of a kilogram
of apples produced in Italy in the European supply chain \citep{Muder2022}. More details are shown in the
SI (Section~2), in figures~S6--S10. The LCIA was performed using the IPCC~2021 method in Brightway2 v2.4.2
\citep{Mutel2017}. This step provides the pipeline anchor information as input for setting the ground for
the LLM interpretation of the LCA findings (Table~\ref{tab:3}). In essence, the LCIA
(\textbf{Figure~\ref{fig:4}}) shows that the main hotspot is diesel consumption, mainly coming from
transport via production (18\%) and usage (59\%). Based on a 50\% reduction target in diesel consumption,
we confirm that emissions would appreciably decrease. Then, as an alternative to diesel, we propose
renewable hydrogen to curb emissions, but the implementation broad technical, socio-political and economic
implications remain unclear. Our tool is then coupled with the LCIA results to shed light on the latter
aspects of the analysis. Overall, the green hydrogen substitution is not proposed by the LCA, but instead,
we use the AI-assisted pipeline to explore this alternative from a high-level multi-perspective viewpoint.

\begin{table}[h]
  \centering
  \caption{LCA practitioner anchor for the LLM as step~5.1, including the case study functional unit and
    main detected bottlenecks from the LCIA; and the timeline for implementation, focus, context and
    objective to propose a roadmap.}
  \label{tab:3}
  \small
  \begin{tabular}{p{3.8cm} p{10.7cm}}
    \toprule
    Case & Apple production in Italy and export to EU countries \\
    \midrule
    Functional unit & 1~kg of apples \\
    \midrule
    Target year & 2030 \\
    \midrule
    Hotspot(s) & High emissions from diesel combustion in:
      \begin{itemize}[nosep,leftmargin=*,topsep=2pt,after=\vspace{2pt}]
        \item Orchard field machinery (tractors, harvesters)
        \item On-site mobile equipment (forklifts, loaders)
        \item Internal logistics vehicles
        \item Outbound transport
      \end{itemize} \\
    \midrule
    Transition target & 50\% reduction in direct diesel use by 2030, without reducing production output \\
    \midrule
    Primary focus & Evaluate green hydrogen substitution for direct diesel combustion uses \\
    \midrule
    Context boundaries &
      \begin{itemize}[nosep,leftmargin=*,topsep=2pt,after=\vspace{2pt}]
        \item Stationary processing equipment and refrigeration are primarily electrified (not assumed
          hydrogen-replaceable unless explicitly evaluated).
        \item Upstream emissions (e.g., nitrogen fertilisers) are relevant but treated as indirect
          decarbonisation pathways.
      \end{itemize} \\
    \midrule
    Decision objective & Identify technically feasible, economically realistic, and policy-aligned
      hydrogen-enabled transition pathways to 2030 \\
    \bottomrule
  \end{tabular}
\end{table}

After providing the pipeline with the base context as a use case anchor, the LLM agents are ready to
receive any specific prompts while the user also configures any related parameters via the GUI environment.
More precisely, we select the number of retrieved context units (vectors) within a range of three to ten.
This range has been only used as an optional configuration parameter. The retrieved context units are
always set to ten across all micro-prompt tests. Subsequently, the perspective is selected from a dropdown
list of categorical values that is responsible for informing the pipeline which vectorised dataspace and
system prompt must be used for responding to the prompt. For evaluating the end-to-end flow of our
pipeline, we utilise a set of 8~prompts consisting of 4~pairs of perspective-based LLM inputs
(Table~\ref{tab:4}). The questions to be asked can undeniably vary from case to case or among
practitioners and researchers. For this reason, we address each perspective with respect to the different
dimensions offered. The public perspective can offer insights and opinions based on the perception and
concerns around costs, safety, and credibility. Further, the business perspective can offer sector-specific
applications, business models, and integration schemes. The academic perspective provides valuable insights
on the technical and systemic feasibility with the corresponding prompts aiming to extract information
focused on these aspects. Finally, the EU funding perspective prompts aim to provide insights on
acceleration mechanisms and sector-specific funding initiatives that could assist the targeted transition
alongside the networking and knowledge transfer. This is a proof-of-concept implementation that could be
further refined and expanded in future works. The full responses of the micro-prompts are included in the
SI (Section~3).

\begin{table}[h]
  \centering
  \caption{Micro-prompts used for perspective evidence-based reasoning in step~5.2.}
  \label{tab:4}
  \small
  \begin{tabular}{clp{2.8cm}p{7.8cm}}
    \toprule
    \textbf{ID} & \textbf{Perspective} & \textbf{Domain} & \textbf{Prompt} \\
    \midrule
    1 & Public    & Legitimacy \& risk
      & How is hydrogen adoption in agricultural production perceived in terms of sustainability credibility
        and green branding? \\[3pt]
    2 & Public    & Legitimacy \& risk
      & What safety, cost, or infrastructure concerns are commonly expressed regarding hydrogen use in
        industrial or agricultural systems? \\[3pt]
    3 & Business  & Business integration
      & What commercially viable models exist for hydrogen procurement or on-site hydrogen production in
        industrial facilities? \\[3pt]
    4 & Business  & Business integration
      & How is hydrogen integrated into supply chains or input sourcing (e.g., fertilisers, logistics) in
        agri-food systems? \\[3pt]
    5 & Academic  & Technical \& system feasibility
      & What are the technical feasibility conditions and efficiency trade-offs of replacing diesel with
        green hydrogen in agricultural machinery and transport? \\[3pt]
    6 & Academic  & Technical \& system feasibility
      & How does integrating green hydrogen across multiple stages of an agricultural production system
        affect lifecycle emissions and system efficiency? \\[3pt]
    7 & EU funding & Acceleration \& governance
      & What funded projects or pilots support hydrogen deployment in agriculture or heavy-duty
        transport? \\[3pt]
    8 & EU funding & Acceleration \& governance
      & What regulatory frameworks or funding instruments could accelerate hydrogen integration across
        agri-food value chains by 2030? \\
    \bottomrule
  \end{tabular}
\end{table}

\section{Results and Discussion}

The findings are presented following the sequential logic of the proposed framework. First, we examine the
perspective-conditioned outputs generated during the micro-prompt stage, summarising the principal evidence
patterns and interpretative insights emerging from the academic, business, public, and EU funding
perspectives. Subsequently, the ledger-constrained outputs were transferred to the neutral synthesis stage,
where a dedicated non-retrieval LLM consolidated the previously grounded evidence under a separate synthesis
prompt, without introducing additional retrieved context. This process produced an extended
implementation-oriented synthesis report capturing transition pathways, enabling mechanisms, critical
constraints, and cross-perspective tensions relevant to the case study. The key findings are summarised
below, while the full synthesis output is included in the SI (Section~5).

\subsection{Micro-prompting (step 5.2)}

This subsection summarises the paired prompt outputs per perspective. GPT-5 nano was used to generate a
concise summary of the key points raised by each perspective LLM agent. More precisely, each subsection
summarises the two prompt responses per perspective rather than the individual retrieved context blocks
(retrieved comments, descriptions, or abstracts), to avoid introducing new context via the model. The full
micro-prompt responses are included in the SI for full reference (Section~3). For further quality testing, we employ LLM-based quality flags for evaluating the generated LLM responses based on the retrieved context (Table~\ref{tab:5}).  

\begin{table}[h]
  \centering
  \caption{Summary of the retrieval quality flags assigned to each micro-prompt response during the
    perspective-conditioned retrieval step~(5.2). The indicators reflect the LLM-based evaluation of
    semantic relevance, within-perspective coverage, evidence density, and overall grounding strength for
    the retrieved context units. Detailed descriptions of the quality flag definitions and interpretation
    criteria are provided in Table~\ref{tab:2}.}
  \label{tab:5}
  \begin{tabular}{cccc}
    \toprule
    \textbf{Prompt ID} & \textbf{Relevance} & \textbf{Coverage} & \textbf{Evidence density} \\
    \midrule
    1 & high     & multi-view & strong \\
    2 & high     & multi-view & strong \\
    3 & high     & multi-view & strong \\
    4 & moderate & multi-view & sparse \\
    5 & high     & multi-view & strong \\
    6 & high     & multi-view & strong \\
    7 & high     & multi-view & strong \\
    8 & high     & multi-view & strong \\
    \bottomrule
  \end{tabular}
\end{table}

\subsubsection{Public: legitimacy, safety, and deployment risk}

Public discourse frames hydrogen adoption in agriculture as a potentially useful but contested
decarbonisation pathway. While hydrogen is frequently promoted by policymakers and industry actors as a
``green'' transition technology, its credibility is often questioned due to concerns about greenwashing and
the expectation that production, storage, and transport costs will remain high in the medium term. In this
context, the suitability of hydrogen-based solutions is perceived to depend strongly on the provenance of
hydrogen~--- specifically whether it is produced using renewable electricity~--- and on transparent
lifecycle accounting rather than branding narratives alone. Safety and infrastructure readiness emerge as
dominant perceived risks, particularly regarding hydrogen storage, transport, and regulatory approval
processes. Public discourse frequently highlights the need for specialised infrastructure, safety controls,
and permitting procedures, while also raising concerns on the uncertainty about the feasibility of
large-scale deployment in sectors that do not already operate hydrogen supply chains. At the same time,
some niche applications~--- such as hydrogen-powered heavy machinery~--- are acknowledged as potentially
practical where operational factors such as refueling time, energy density, or grid limitations make
electrification difficult. Overall, the public perspective suggests that hydrogen integration into
agricultural systems will require demonstrable safety, credible renewable sourcing, and clear operational
advantages to overcome skepticism and support its role in reducing diesel-related emissions identified in
the LCA hotspot analysis.

\subsubsection{Business: business model integration and value chain structuring}

From the industry perspective, hydrogen adoption in agricultural production systems is primarily framed as
a procurement and infrastructure integration challenge rather than a purely technological substitution.
Commercially viable models already emerging in the hydrogen economy include on-site hydrogen generation
through leasing or managed-service models, decentralised small-scale production near industrial users,
mobile or autonomous refueling infrastructure, and conventional procurement from large-scale hydrogen
supply projects. These models emphasise reducing upfront capital requirements and enabling flexible supply
arrangements, often supported by procurement marketplaces or contractor platforms that simplify supplier
discovery and tendering processes. Such service-oriented deployment strategies may lower the barriers for
facilities seeking to reduce diesel consumption in mobile machinery, internal logistics, or transport
operations identified as LCA hotspots.

However, direct evidence of hydrogen integration within agricultural input supply chains~--- particularly
fertiliser production or farm-level energy systems~--- remains limited in the retrieved business corpus, a
finding that was corroborated by the sparse evidence density flag returned for the corresponding
micro-prompt; the only such instance across all eight prompts. This diagnostic signal suggests that
hydrogen's role in agri-food input supply chains remains an underrepresented topic in current business
discourse, pointing to a meaningful gap for future data collection and query refinement. As a result,
near-term integration pathways appear more mature in logistics and energy supply infrastructure than in
upstream agri-food inputs, suggesting that early hydrogen deployment may occur through service-based fuel
supply and mobility solutions rather than full value-chain transformation.

\subsubsection{Academic: technical feasibility and system-level trade-offs}

The academic literature frames hydrogen substitution for diesel in agricultural machinery and transport as
technically plausible but highly conditional. Feasibility depends primarily on the availability of a
reliable and cost-competitive hydrogen supply chain, including production, transport, storage, and
site-level demand sufficient to justify infrastructure investments. Studies indicate that on-site hydrogen
production becomes economically attractive only under favorable electricity prices and adequate demand
scale, while logistics choices such as liquid hydrogen, ammonia carriers, or centralised production
significantly affect cost and deliverability. Direct replacement of diesel engines requires dedicated
hydrogen combustion or fuel-cell technologies, as partial blending with existing diesel systems presents
operational and performance challenges. From a systems perspective, lifecycle emissions and efficiency
gains depend strongly on the electricity mix used for electrolysis and the energy losses associated with
hydrogen production, storage, and distribution. These thermodynamic and infrastructure constraints mean
that hydrogen does not automatically outperform alternative decarbonisation pathways, particularly where
electrification is feasible. Moreover, the literature highlights a lack of agriculture-specific empirical
studies, implying that projections for orchard machinery or farm logistics must currently rely on
extrapolation from other industrial or transport sectors.

\subsubsection{EU funding: institutional acceleration and demonstration ecosystems}

The EU innovation and funding landscape frames hydrogen deployment primarily through large-scale
demonstrator programs and regulatory mechanisms designed to accelerate infrastructure development and
technology readiness. Existing projects largely focus on heavy-duty transport and regional hydrogen
ecosystems, including fleet demonstrations, hydrogen refueling networks, and integrated hydrogen valleys
that connect mobility, energy production, and industrial users. These initiatives provide operational
evidence for hydrogen-powered vehicles and logistics infrastructure, offering potential indirect pathways
for reducing diesel use in agricultural supply chains~--- particularly in transport and distribution stages
linked to agri-food production. In parallel, EU funding instruments and regulatory frameworks aim to
accelerate hydrogen deployment through innovation funding, test-bed programs, ecosystem planning tools,
and harmonised safety and fuel-quality standards. These mechanisms are intended to reduce regulatory
barriers, support pilot deployments, and enable cross-regional replication of hydrogen solutions by 2030.
However, within the retrieved policy corpus, explicit agriculture-focused hydrogen pilots remain limited,
indicating that most institutional momentum currently targets mobility and energy infrastructure rather
than on-farm machinery or agricultural production systems.

\subsection{Neutral synthesis (step 5.3)}

\begin{figure}[h]
  \centering
  \includegraphics[width=\textwidth]{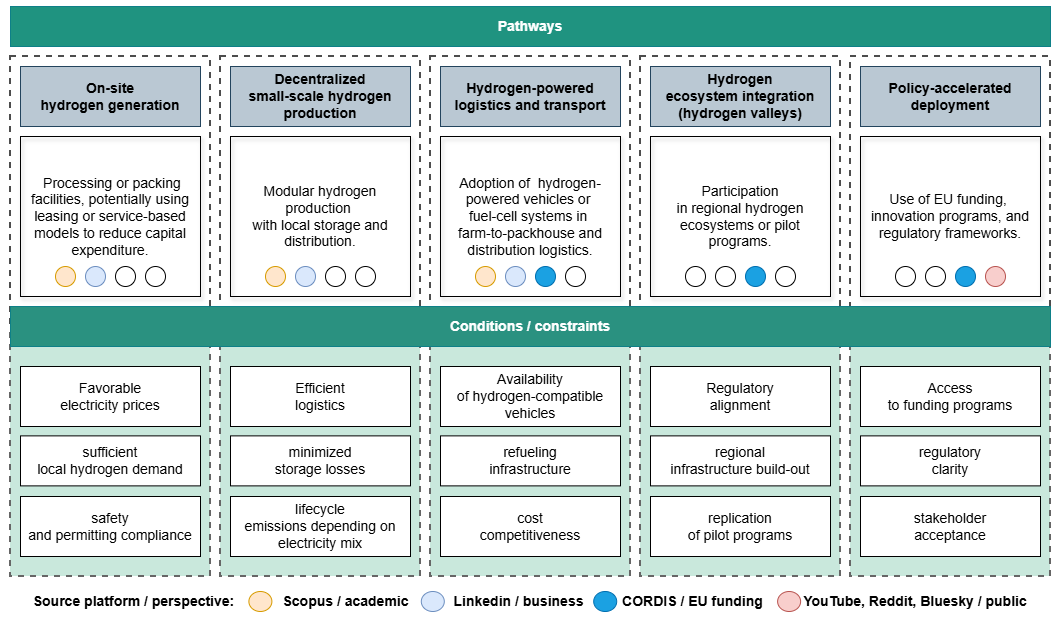}
  \caption{Ledger-constrained 2030 green hydrogen transition roadmap for Italian apple production,
    synthesised from perspective-conditioned retrieval outputs. The upper panel presents five deployment
    pathways and the lower panel maps their corresponding conditions and constraints. Circle indicators
    denote the contributing source perspective (Scopus/academic, LinkedIn/business, CORDIS/EU funding,
    YouTube--Reddit--Bluesky/public) for each pathway-constraint pair.}
  \label{fig:5}
\end{figure}

In this subsection we elaborate further on the final perspective synthesis of the micro-prompts
(step~5.3). The detailed version of the synthesis LLM output is included in the SI (Section~5). In this
context, we first ask the LLM (no retrieval) to provide the most effective pathways and underlying
conditions and constraints that can affect the green hydrogen integration to our system
(\textbf{Figure~\ref{fig:5}}). Then, it summarises all the micro-prompt outputs with emphasis given in
highlighting the different points of view alongside the convergence areas. The perspective focus is
affected from the micro-prompt design and context; however, it effectively showcases the presence or
absence of specific evidence in the given discourse.

\subsubsection{Pathways and caveats}

The synthesis indicates that achieving the 50\% diesel reduction target by 2030 is most plausible through
a staged combination of the following pathways rather than relying on a single deployment model. For
on-site generation, the analysis suggests focusing on localised electrolysis at processing facilities,
utilising leasing or service-based models to circumvent high initial capital barriers. Further, the
decentralised production pathway emphasises modular units near farm clusters to provide local storage and
reduce distribution-related energy losses. The direct substitution of diesel is most prominently addressed
through the logistics and transport pathway, which targets fuel-cell adoption for farm-to-packhouse
distribution. To support these technical shifts, the ecosystem integration pathway advocates for
participation in regional hydrogen valleys to synchronise production and storage across industrial sectors,
while the policy-accelerated pathway identifies the strategic use of EU funding and regulatory frameworks
to de-risk these transitions across the entire value chain. These pathways are shaped through the lenses
of the four stakeholder agents, concluding that the best plan forward would consist of a hybrid approach
based on integrating modular on-site production with fuel-cell-based logistics.

\subsubsection{Perspective convergence and uncertainties}

Academic evidence highlights that hydrogen substitution potential is conditioned by supply-chain maturity,
electricity sourcing for electrolysis, and efficiency losses associated with hydrogen production, storage,
and transport. From a business perspective, emerging deployment models, including on-site hydrogen
generation services, decentralised production systems, and hydrogen procurement platforms, indicate that
commercially viable supply pathways are developing, particularly where capital risk can be reduced through
service-based arrangements. Public discourse introduces additional constraints related to safety
perceptions, infrastructure readiness, and skepticism toward green branding, emphasising the importance of
transparent lifecycle performance and credible renewable sourcing. At the institutional level, EU
innovation programs and hydrogen valley initiatives demonstrate substantial momentum in mobility and
infrastructure deployment, although agriculture-specific hydrogen pilots remain limited in the retrieved
evidence base. Taken together, the perspectives converge on a staged transition strategy combining
decentralised or on-site hydrogen production, hydrogen-enabled logistics and transport applications, and
participation in regional hydrogen ecosystems supported by EU funding and regulatory frameworks. However,
the synthesis also highlights key uncertainties related to lifecycle emissions, economic viability, and the
lack of sector-specific LCA data for agricultural hydrogen applications, indicating that pilot-based
deployment and monitoring will be essential to validate the feasibility of reaching the 2030 diesel
reduction target.

Across perspectives, hydrogen is viewed as technically feasible but strongly context-dependent, with
deployment outcomes shaped by infrastructure maturity, policy support, and supply-chain economics.
Although we did not use the same micro-prompts across different perspectives, there have been several key
points raised as cross-domain tensions and convergences. Most importantly, convergences refer to the
hydrogen viability but also the conditional aspects of it. The academic and hydrogen viability but also the conditional aspects of it. The academic and business perspectives both
touch this narrative. Further, the infrastructure maturity is central as an enabler, which was discussed
in the public text corpus and also described in the EU project objectives. The policy support is
significant to the green hydrogen adoption but mobility-focused, according to the EU funding initiatives.
On the other hand, tensions encapsulate the high-level view of potential mismatches. Business optimism
mainly stems from organisations' needs to attract customers or investors while the academic world adopts a
more conservative approach regarding the implementation of real green and sustainable hydrogen solutions.
Public opinion remains suspicious and raises skepticism about the actual feasibility, cost, and widespread
adoption. On the contrary, a wide policy-level hydrogen promotion is observed. In terms of funding, strong
mobility pilots are supported but there are limited agriculture-specific direct applications. On top of
these, there is a set of potential implications or critical aspects to be considered during the
implementation pathway (Table~\ref{tab:6}).

\begin{table}[h]
  \centering
  \caption{Key uncertainties and constraints affecting hydrogen deployment within the proposed analytical
    extension.}
  \label{tab:6}
  \small
  \begin{tabular}{p{4.2cm} p{3.2cm} p{6.6cm}}
    \toprule
    \textbf{Issue} & \textbf{Source perspective} & \textbf{Use case implications} \\
    \midrule
    Lifecycle emissions uncertainty
      & Academic
      & Net climate benefits depend on renewable electricity sourcing and hydrogen logistics efficiency. \\[3pt]
    Economic viability
      & Academic, Business
      & Hydrogen deployment may require favorable electricity prices, sufficient demand scale, and policy
        incentives. \\[3pt]
    Safety and infrastructure readiness
      & Public
      & Storage, transport, and regulatory approval requirements could slow adoption without robust safety
        frameworks. \\[3pt]
    Evidence gap for agriculture-specific hydrogen systems
      & Academic
      & Limited empirical LCA data for orchard machinery and farm operations requires pilot-based
        validation. \\[3pt]
    Regulatory and market conditions
      & EU innovation / funding
      & Deployment timelines depend on regulatory harmonisation and access to EU funding mechanisms. \\
    \bottomrule
  \end{tabular}
\end{table}

Perspective-based retrieval supports interpretative rigor in LCA by transforming the traditionally
``black-box'' interpretation phase into a novel implementation-oriented analytical layer that integrates
technical, market, policy, and societal evidence. The data decoupling into distinct academic, industry,
public, and policy lenses ensures that the translation of quantified hotspots, such as the diesel emissions
identified in the Italian apple production, is grounded in a balanced synthesis of technical feasibility,
market readiness, and societal legitimacy. This modular architecture is designed to structurally reduce
hallucination risk by enforcing constrained retrieval from heterogeneous corpora and utilising a
ledger-based system that limits the introduction of ungrounded claims during final synthesis. Ultimately,
this approach allows us to move beyond simple impact assessment toward a disciplined extension of LCA
established phases, providing decision-makers with traceable 2030 decarbonisation pathways that account
for real-world constraints like infrastructure gaps, funding accelerators, and stakeholder skepticism.

\section{Conclusions}

Here we propose an RAG-based approach to enhance the interpretation phase of LCA, which provides valuable
insights regarding environmental improvement opportunities and their adoption at scale. The proposed
analytical pipeline, framed as a web-based RAG application, demonstrates the feasibility of an AI-assisted
implementation-oriented extension to conventional LCA interpretation. While presented here as a
proof-of-concept, future expansion through broader datasets, benchmarking, and methodological validation
could further support LCA practice. Through this perspective-conditioned architecture, the framework
effectively translates quantified LCA hotspots~--- such as the high diesel emissions identified in Italian
apple production~--- into actionable, multi-dimensional strategic roadmaps. The study reveals that while
technical feasibility for green hydrogen substitution is grounded in academic literature, its successful
deployment is contingent upon navigating the complexities of public legitimacy, industry-service models,
and EU-level funding accelerators. Ultimately, the integration of structured retrieval with constrained
synthesis provides an auditable, evidence-based foundation for decision-makers to move beyond impact
assessment toward tangible decarbonization.

Future research will focus on enhancing the system's data scalability by integrating a broader array of
document types, including ISO~14040/44 standards, national and EU legislation, and corporate sustainability
reports. A critical next step involves the deployment of the application on a local server using local
LLMs (e.g., Llama-3 or Mistral) to ensure that sensitive or proprietary corporate data are never exposed
via external APIs. This local-first approach will facilitate the secure ingestion of private data and
internal surveys, allowing for highly tailored organizational roadmaps.

\section*{Acknowledgements}
This publication was created as part of NCCR Catalysis (grant
number 225147), a National Centre of Competence in Research
funded by the Swiss National Science Foundation.

\section*{Data availability statement}
The datasets generated and analysed during this study cannot be made publicly available due to several constraints inherent to the data collection process. The academic corpus was retrieved via the Elsevier Scopus API under institutional access credentials, and redistribution of the raw abstracts is restricted by Elsevier's terms of service. The industry corpus was collected from LinkedIn company profiles using a web scraping software tool; redistribution of this data is prohibited under LinkedIn's terms of use. The public discourse corpus was gathered via the YouTube Data API v3, Reddit API, and Bluesky API using registered developer keys, and the underlying content contains usernames and user-generated text that may constitute personally identifiable information, precluding open sharing. The EU funding corpus was sourced from the CORDIS platform, which requires an EU login for bulk data access, and full redistribution falls outside the scope of permitted reuse. The LCA background data used for the case study were obtained from the ecoinvent database v3.12, which is available under a commercial licence from the ecoinvent Association (www.ecoinvent.org). The foreground LCA data and LCIA results generated during the study are available from the corresponding author upon reasonable request.



\begin{thebibliography}{99}
 
\bibitem[Abo El-Enen et al.(2025)]{AboElEnen2025}
Abo El-Enen M, Saad S, Nazmy T (2025) A survey on retrieval-augmentation generation (RAG) models for
healthcare applications. Neural Comput Appl 37:28191--28267.
\url{https://doi.org/10.1007/s00521-025-11666-9}
 
\bibitem[Algunaibet and Guill\'{e}n-Gos\'{a}lbez(2019)]{Algunaibet2019}
Algunaibet IM, Guill\'{e}n-Gos\'{a}lbez G (2019) Life cycle burden-shifting in energy systems designed
to minimise greenhouse gas emissions. J Clean Prod 229:886--901.
\url{https://doi.org/10.1016/j.jclepro.2019.04.276}
 
\bibitem[Arslan et al.(2024)]{Arslan2024}
Arslan M, Munawar S, Cruz C (2024) Business insights using RAG--LLMs: a review and case study. J Decis
Syst 1--30. \url{https://doi.org/10.1080/12460125.2024.2410040}
 
\bibitem[Azapagic(1999)]{Azapagic1999}
Azapagic A (1999) Life cycle assessment and its application to process selection, design and
optimisation. Chemical Engineering Journal 73:1--21.
\url{https://doi.org/10.1016/S1385-8947(99)00042-X}
 
\bibitem[Balaguer et al.(2024)]{Balaguer2024}
Balaguer A, Benara V, Cunha RL de F, et al (2024) RAG vs Fine-tuning: Pipelines, Tradeoffs, and a Case
Study on Agriculture.
 
\bibitem[Bang et al.(2025)]{Bang2025}
Bang Y, Ji Z, Schelten A, et al (2025) HalluLens: LLM Hallucination Benchmark. In: Proceedings of the
63rd Annual Meeting of the Association for Computational Linguistics. pp 24128--24156.
 
\bibitem[Douze et al.(2026)]{Douze2026}
Douze M, Guzhva A, Deng C, et al (2026) The Faiss Library. IEEE Trans Big Data 12:346--361.
\url{https://doi.org/10.1109/TBDATA.2025.3618474}
 
\bibitem[Goridkov et al.(2024)]{Goridkov2024}
Goridkov N, Wang Y, Goucher-Lambert K (2024) What's in this LCA Report? Procedia CIRP 122:964--969.
\url{https://doi.org/10.1016/J.PROCIR.2024.01.131}
 
\bibitem[Hasan et al.(2026)]{Hasan2026}
Hasan MdT, Waseem M, Kemell K-K, et al (2026) Engineering RAG Systems for Real-World Applications.
pp 143--158.
 
\bibitem[ISO(2014)]{ISO14044}
ISO 14044 (2014) Environmental management -- Life cycle assessment -- Requirement and guidelines.
 
\bibitem[J et al.(2024)]{J2024}
J~MR, VM~K, Warrier H, Gupta Y (2024) Fine Tuning LLM for Enterprise: Practical Guidelines and
Recommendations.
 
\bibitem[Ji et al.(2023)]{Ji2023}
Ji Z, Yu T, Xu Y, et al (2023) Towards Mitigating LLM Hallucination via Self Reflection. In: Findings
of the Association for Computational Linguistics: EMNLP~2023. pp 1827--1843.
 
\bibitem[Keoleian(1993)]{Keoleian1993}
Keoleian GA (1993) The application of life cycle assessment to design. J Clean Prod 1:143--149.
\url{https://doi.org/10.1016/0959-6526(93)90004-U}
 
\bibitem[Khorasani et al.(2025)]{Khorasani2025}
Khorasani M, Abdou M, Hern\'{a}ndez Fern\'{a}ndez J (2025) Streamlit Basics. In: Streamlit for Web
Development. Apress, Berkeley, CA, pp 31--66.

\bibitem[Kumar et al.(2026)]{Kumar2026}
Kumar A, Kim S, Bakshi B. R. (2026). Role of artificial intelligence in the chemical industry transition to a sustainable, circular, and net-zero future. Current Opinion in Chemical Engineering, 51, 101234. 
\url{https://doi.org/10.1016/j.coche.2026.101234}
 
\bibitem[Lewandowska et al.(2011)]{Lewandowska2011}
Lewandowska A, Matuszak-Flejszman A, Joachimiak K, Ciroth A (2011) Environmental life cycle assessment
as a tool for identification of environmental aspects in EMS. Int J Life Cycle Assess 16:247--257.
\url{https://doi.org/10.1007/s11367-011-0252-3}
 
\bibitem[Lewis et al.(2021)]{Lewis2021}
Lewis P, Perez E, Piktus A, et al (2021) Retrieval-Augmented Generation for Knowledge-Intensive NLP
Tasks.
 
\bibitem[Lin et al.(2024)]{Lin2024}
Lin X, Wang W, Li Y, et al (2024) Data-efficient Fine-tuning for LLM-based Recommendation. In:
Proceedings of the 47th ACM SIGIR Conference. ACM, New York, NY, USA, pp 365--374.
 
\bibitem[Luan and Jing(2026)]{Luan2026}
Luan K, Jing Y (2026) AI-enabled integration of life cycle assessment in building retrofitting.
Results in Engineering 29:109568. \url{https://doi.org/10.1016/j.rineng.2026.109568}
 
\bibitem[Mohammadi Kashka et al.(2023)]{Mohammadi2023}
Mohammadi Kashka F, Tahmasebi Sarvestani Z, Pirdashti H, et al (2023) Sustainable Systems Engineering
Using LCA: Application of AI for Predicting Agro-Environmental Footprint. Sustainability 15:6326.
\url{https://doi.org/10.3390/su15076326}
 
\bibitem[Muder et al.(2022)]{Muder2022}
Muder A, Garming H, Dreisiebner-Lanz S, et al (2022) Apple production and apple value chains in Europe.
Eur J Hortic Sci 87:1--22. \url{https://doi.org/10.17660/eJHS.2022/059}
 
\bibitem[Mutel(2017)]{Mutel2017}
Mutel C (2017) Brightway: An open source framework for Life Cycle Assessment. J Open Source Softw
2:236. \url{https://doi.org/10.21105/joss.00236}
 
\bibitem[Neha et al.(2025)]{Neha2025}
Neha F, Bhati D, Shukla DK (2025) Retrieval-Augmented Generation (RAG) in Healthcare: A Comprehensive
Review. AI 6:226. \url{https://doi.org/10.3390/ai6090226}
 
\bibitem[Nwagwu et al.(2025)]{Nwagwu2025}
Nwagwu CC, Ogorodnyk O, S\o{}lvsberg E, et al (2025) Integrating Artificial Intelligence into Life
Cycle Assessment. Journal of Sustainable Metallurgy 11:3590--3605.
\url{https://doi.org/10.1007/s40831-025-01305-x}
 
\bibitem[Petrosa et al.(2025)]{Petrosa2025}
Petrosa D, Haverkamp P, Backes JG, et al (2025) Development of a BIM-based AI-driven matching tool
for LCA datasets. Discover Sustainability 6:1237.
\url{https://doi.org/10.1007/s43621-025-02203-8}
 
\bibitem[Prado et al.(2022)]{Prado2022}
Prado V, Seager TP, Guglielmi G (2022) Challenges and risks when communicating comparative LCA results
to management. Int J Life Cycle Assess 27:1164--1169.
\url{https://doi.org/10.1007/s11367-022-02090-5}

\bibitem[Preuss et al.(2024)]{Preuss2024}
Preuss N, Alshehri A. S., You F. (2024). Large language models for life cycle assessments: Opportunities, challenges, and risks. Journal of Cleaner Production, 466, 142824.
\url{https://doi.org/10.1016/j.jclepro.2024.142824}
 
\bibitem[Preuss and You(2026)]{Preuss2026}
Preuss N, You F (2026) Automating Life Cycle Assessments through Artificial Intelligence Agents.
Environ Sci Technol 60:33--48. \url{https://doi.org/10.1021/acs.est.5c14493}
 
\bibitem[Rawte et al.(2023)]{Rawte2023}
Rawte V, Chakraborty S, Pathak A, et al (2023) The Troubling Emergence of Hallucination in Large
Language Models. In: Proceedings of EMNLP 2023. pp 2541--2573.
 
\bibitem[Ren and Sutherland(2025)]{Ren2025}
Ren Y, Sutherland DJ (2025) Learning Dynamics of LLM Finetuning.
 
\bibitem[Ross and Evans(2002)]{Ross2002}
Ross S, Evans D (2002) Use of Life Cycle Assessment in Environmental Management. Environ Manage
29:132--142. \url{https://doi.org/10.1007/s00267-001-0046-7}
 
\bibitem[\v{S}ekrst(2025)]{Sekrst2025}
\v{S}ekrst K (2025) Hallucinations. In: The Illusion Engine. Springer Nature Switzerland, Cham,
pp 211--226.
 
\bibitem[Shafiq et al.(2024)]{Shafiq2024}
Shafiq M, Ayub S, Muthevi A~kumar, Prabhu MR (2024) AI-driven Life Cycle Assessment for sustainable
hybrid manufacturing and remanufacturing. Int J Adv Manuf Technol.
\url{https://doi.org/10.1007/s00170-024-14930-9}
 
\bibitem[Swacha and Gracel(2025)]{Swacha2025}
Swacha J, Gracel M (2025) Retrieval-Augmented Generation (RAG) Chatbots for Education: A Survey.
Applied Sciences 15:4234. \url{https://doi.org/10.3390/app15084234}
 
\bibitem[Verrier et al.(2022)]{Verrier2022}
Verrier B, Li P-H, Pye S, Strachan N (2022) Incorporating social mechanisms in energy decarbonisation
modelling. Environ Innov Soc Transit 45:154--169.
\url{https://doi.org/10.1016/j.eist.2022.10.003}
 
\bibitem[Wernet et al.(2016)]{Wernet2016}
Wernet G, Bauer C, Steubing B, et al (2016) The ecoinvent database version~3 (part~I): overview and
methodology. Int J Life Cycle Assess 21:1218--1230.
\url{https://doi.org/10.1007/s11367-016-1087-8}
 
\bibitem[Yu et al.(2025)]{Yu2025}
Yu H, Gan A, Zhang K, et al (2025) Evaluation of Retrieval-Augmented Generation: A Survey.
pp 102--120.
 
\bibitem[Zhang(2025)]{ZhangJ2025}
Zhang J (2025) Small Language Models Offer Significant Potential for Science Community.
 
\bibitem[Zhang et al.(2025)]{ZhangX2025}
Zhang X, Guo X, Zhao J, et al (2025) Intelligent application of large language model to life cycle
assessment methodology. J Clean Prod 529:146776.
\url{https://doi.org/10.1016/J.JCLEPRO.2025.146776}
 
\end{thebibliography}
\end{document}


\graphicspath{{Figures SI/}}

\maketitle
\thispagestyle{empty}

\section{Dataset description per perspective}

This section provides more detail on the descriptive aspects of the datasets collected from diverse online
platforms. Namely, we collected data from Scopus, LinkedIn, CORDIS, YouTube, Reddit, and Bluesky. These
platforms served as perspective data pools for collecting information related to hydrogen with more focus on
green and renewable hydrogen and feedstocks, where available. Thus, the main perspectives considered are
the Academic (Scopus), Business (LinkedIn), EU funding (CORDIS), and Public (YouTube, Reddit, and
Bluesky).

\subsection{Academic}

\begin{figure}[H]
  \centering
  \begin{subfigure}{\textwidth}
    \centering
    \includegraphics[width=0.82\textwidth]{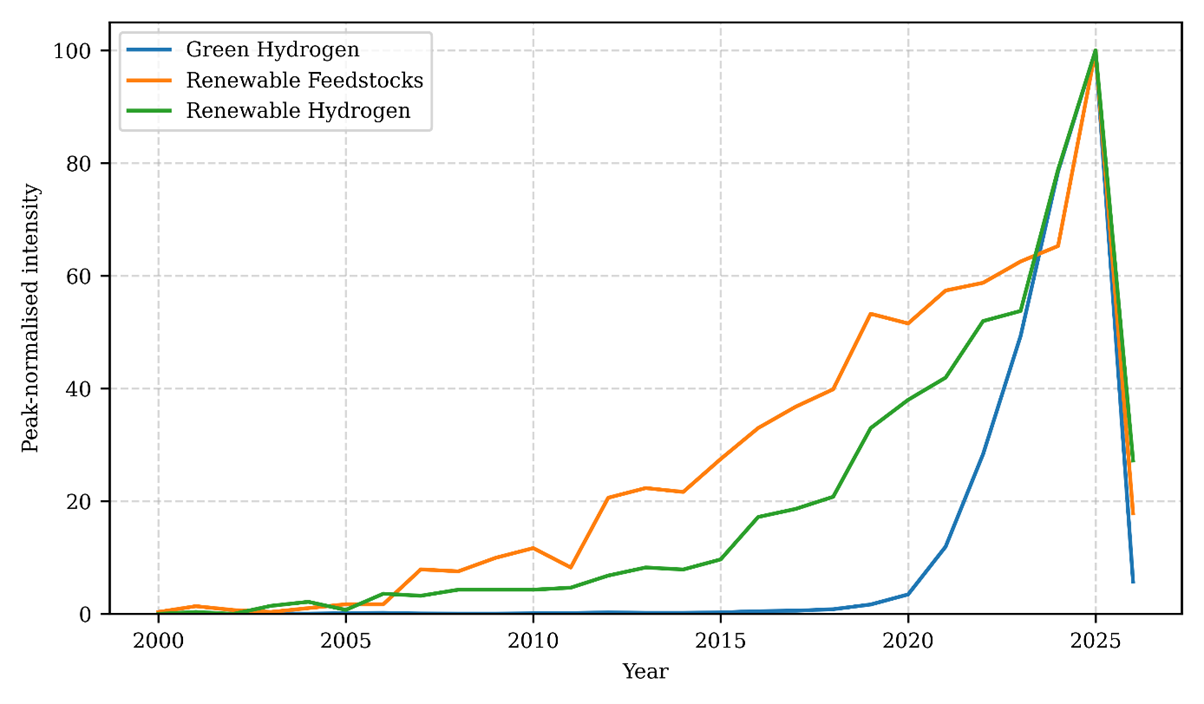}
    \caption{}
  \end{subfigure}\\[8pt]
  \begin{subfigure}{\textwidth}
    \centering
    \includegraphics[width=0.82\textwidth]{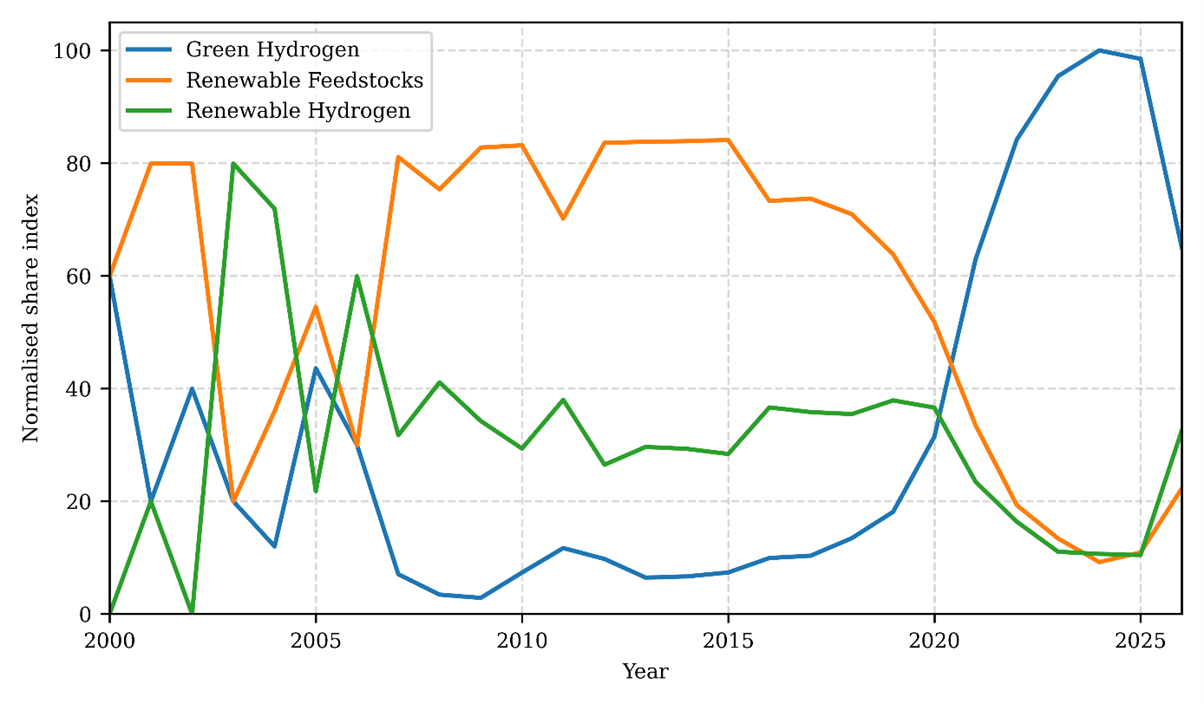}
    \caption{}
  \end{subfigure}
  \caption{\textbf{(a)} Normalised annual mentions of ``green hydrogen,'' ``renewable hydrogen,'' and
    ``renewable feedstock(s)'' in Scopus-indexed literature (2000--2026).
    \textbf{(b)} Relative share of literature index for ``green hydrogen,'' ``renewable hydrogen,'' and
    ``renewable feedstock(s)'' (2000--2026). Values are indexed to the historical maximum share
    ($\max(S)=100$) for visual clarity across datasets; consequently, annual totals exceed 100.}
  \label{fig:S1}
\end{figure}

Scholarly trends from Scopus confirm a definitive shift in terminology and focus. First, while ``renewable
feedstock'' dominated the early relative literature share, ``green hydrogen'' has achieved conceptual
dominance since 2020, with absolute publication volumes following an exponential growth curve that peaked
in 2023--2024. Figure~\ref{fig:S1} illustrates the longitudinal growth in the absolute frequency of mentions
for ``green hydrogen,'' ``renewable hydrogen,'' and ``renewable feedstock(s)'' within the Scopus database
from 2000 to 2026. The data, normalised to a 0--100 scale based on peak mention volume, reveals a period
of relative dormancy followed by a synchronised exponential increase beginning around 2015
(Figure~\ref{fig:S1}a). This trajectory signifies a massive expansion in the total body of literature
dedicated to these fields, with ``green hydrogen'' and ``renewable hydrogen'' reaching their highest absolute
research output between 2022 and 2024. Further, Figure~\ref{fig:S1}b depicts the relative normalised share
of literature as a descriptor of each term's contribution in the total academic discourse.

To facilitate a synchronised comparison across heterogeneous datasets, two distinct normalisation layers
were applied. In these models, the variable $x$ represents the specific entity under analysis (a keyword in
the Scopus dataset or a platform in the YouTube, Reddit, and Bluesky datasets) at a given time $t$.
Figure~\ref{fig:S1}a and Figure~\ref{fig:S5}a utilise a peak-normalisation index to visualise the internal
growth cycle of each entity. Formula~(\ref{eq:norm}) scales the annual frequency against the historical peak
of that specific entity.

\begin{equation}
  \mathrm{norm}_{x,t} = \frac{f(x,t)}{\max(f_x)} \times 100
  \label{eq:norm}
\end{equation}

Figure~\ref{fig:S1}b and Figure~\ref{fig:S5}b illustrate the competitive dominance of an entity within the
total discourse of a specific year. We first calculate the annual share $S$ for both literature and public
discourse (\ref{eq:share}). This identifies the slice of the pie held by a platform or keyword relative to all
other observed entities at that exact moment in time. For maintaining visual consistency and a 0--100
comparison across different datasets, the annual shares are further scaled into a normalised share index
$S_{\mathrm{norm}}$ (\ref{eq:snorm}).

\begin{equation}
  S(x,t) = \frac{f(x,t)}{\sum f(x,t)} \times 100
  \label{eq:share}
\end{equation}

\begin{equation}
  S_{\mathrm{norm}}(x,t) = \frac{S(x,t)}{\max(S)} \times 100
  \label{eq:snorm}
\end{equation}

\subsection{Business}

\begin{figure}[H]
  \centering
  \begin{subfigure}{\textwidth}
    \centering
    \includegraphics[width=0.72\textwidth]{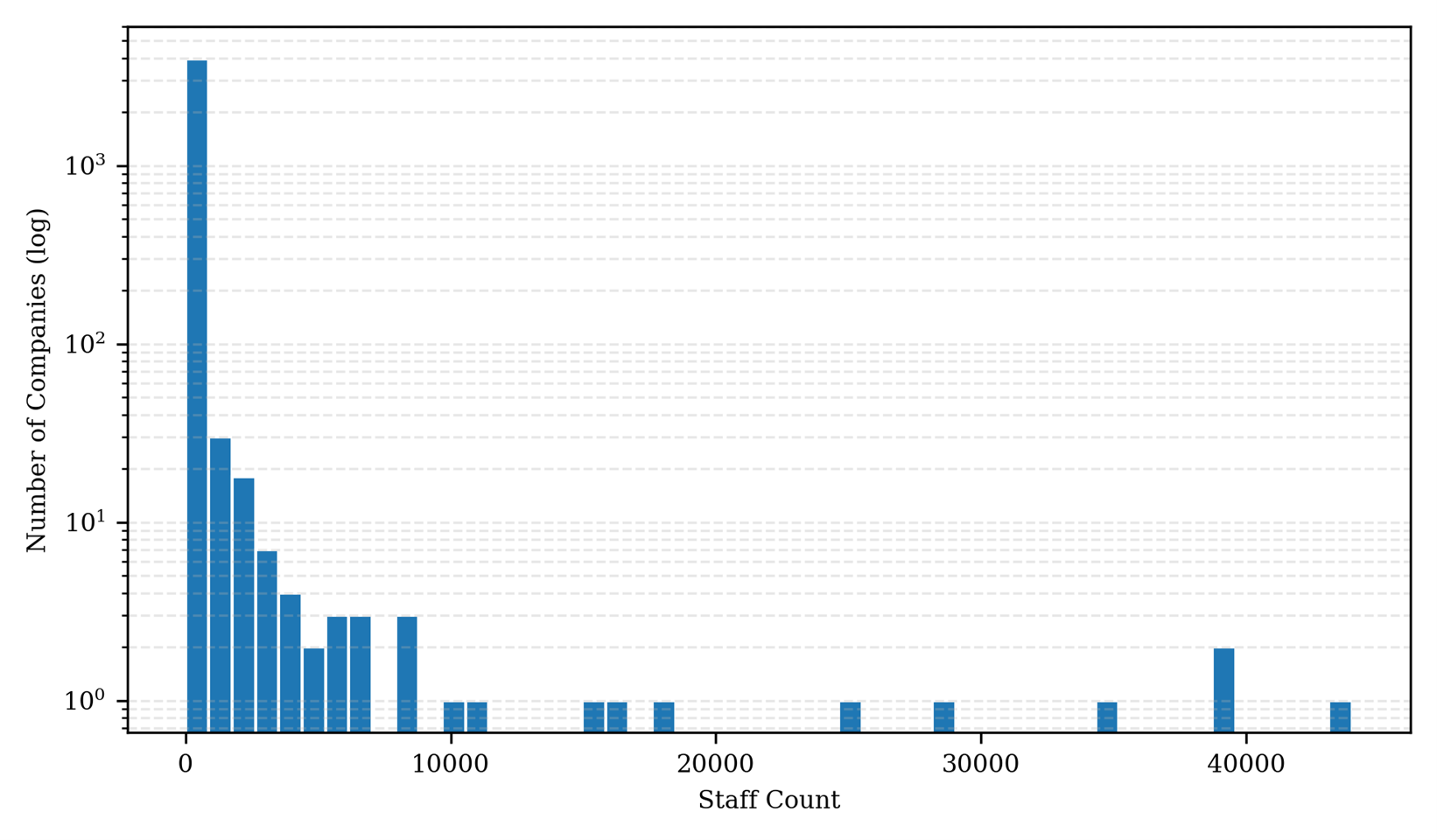}
    \caption{}
  \end{subfigure}\\[6pt]
  \begin{subfigure}{\textwidth}
    \centering
    \includegraphics[width=0.72\textwidth]{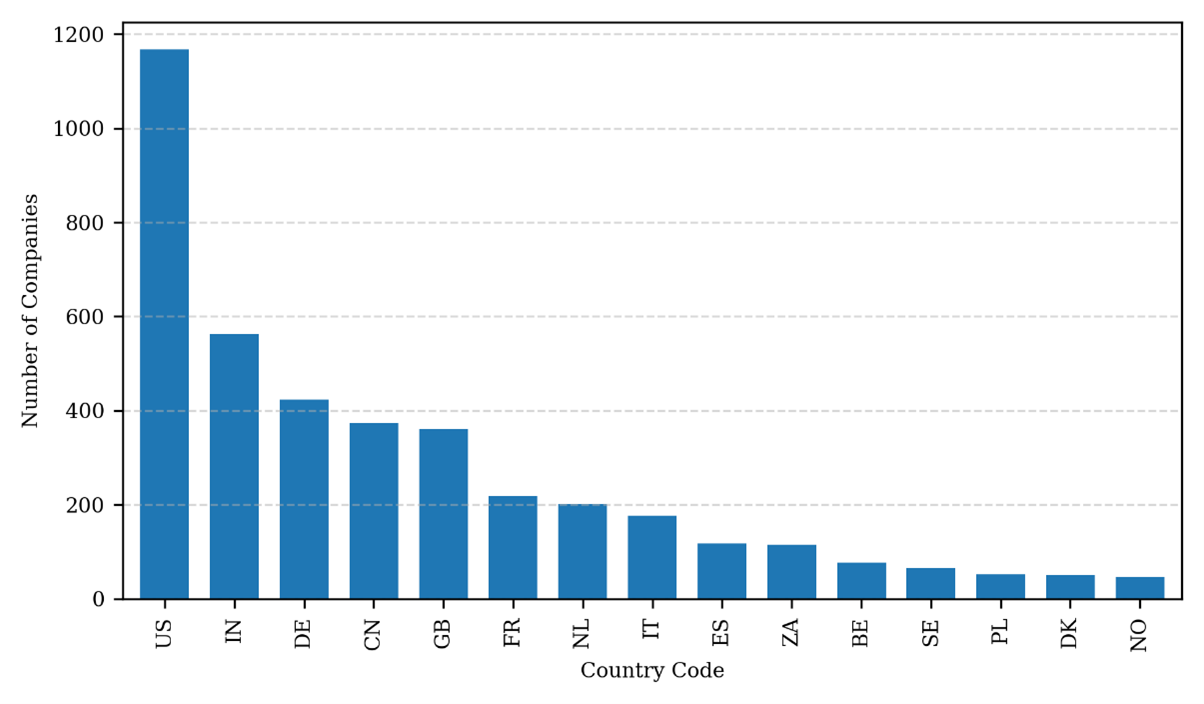}
    \caption{}
  \end{subfigure}\\[6pt]
  \begin{subfigure}{\textwidth}
    \centering
    \includegraphics[width=0.72\textwidth]{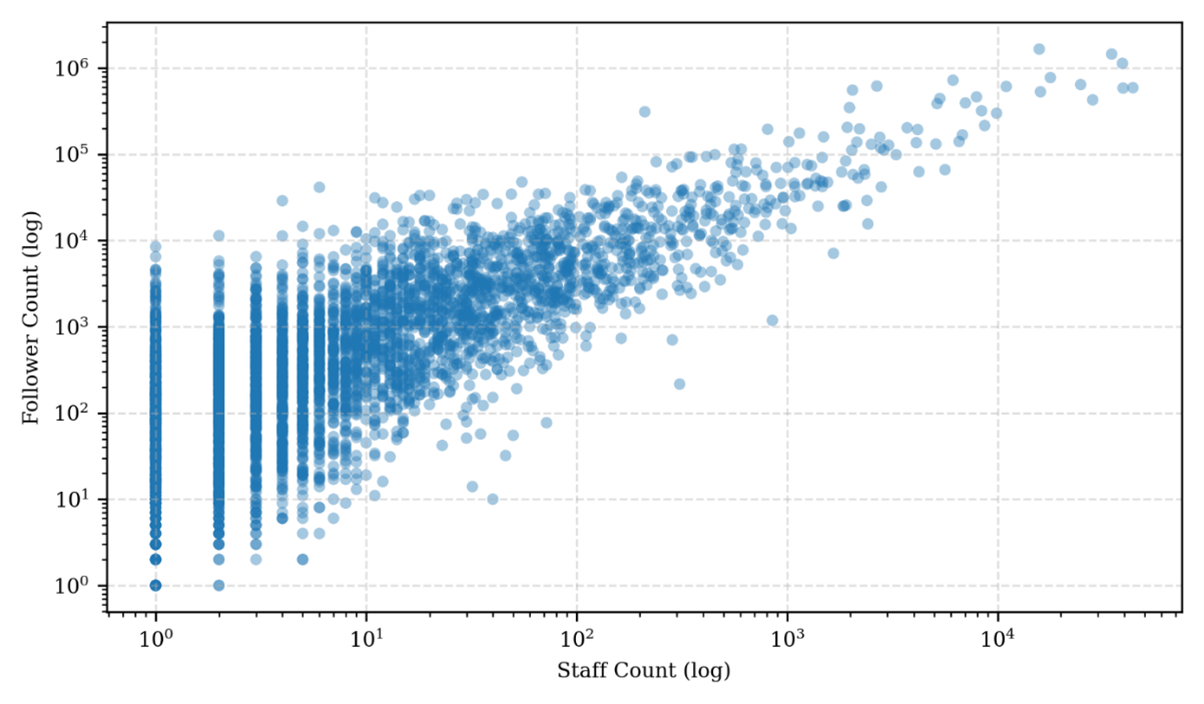}
    \caption{}
  \end{subfigure}
  \caption{Global distribution and digital reach of hydrogen-related companies on LinkedIn.
    \textbf{(a)} Distribution of company size: the frequency of firms by staff count.
    \textbf{(b)} Top 15 headquarter countries: the geographic concentration of hydrogen-related corporate
    profiles.
    \textbf{(c)} Company size vs.\ LinkedIn reach: associated staff and follower count correlation.}
  \label{fig:S2}
\end{figure}

The corporate landscape of the hydrogen sector on LinkedIn is defined by a high density of
small-to-medium enterprises concentrated in a few key geographic hubs. Figure~\ref{fig:S2}a displays the
distribution of company sizes, revealing a heavily right-skewed population where the vast majority of firms
maintain fewer than 1{,}000 employees, despite a long tail of large-scale industrial entities. This
organizational diversity is geographically stratified, as shown in Figure~\ref{fig:S2}b; the United States,
India, and Germany lead the top 15 headquarters countries, indicating that while the hydrogen economy is
global, its digital corporate presence is currently anchored in these primary innovation hubs. The
relationship between organizational scale and digital visibility is characterised by a positive, yet
heterogeneous, correlation. Figure~\ref{fig:S2}c illustrates that while increasing headcount generally
drives higher follower counts, significant dispersion exists among micro-entities. The distinct vertical
stripe-shaped patterns observed at the lower end of the $x$-axis represent the high frequency of
organisations with identical, small integer headcounts (e.g., 1--10 associated members). This visualisation
highlights that in the early-stage hydrogen sector, brand positioning and project novelty allow specialised
technology firms to achieve digital influence that is partially decoupled from their physical staff size.

\begin{table}[H]
  \centering
  \caption{LinkedIn searches used for the business dataset.}
  \label{tab:S1}
  \begin{tabular}{ll}
    \toprule
    \textbf{Industry} & \textbf{Country} \\
    \midrule
    \multirow{6}{*}{Manufacturing}
      & European Union \\
      & United States \\
      & China, India, Iraq, Iran, Russia \\
      & Saudi Arabia, Japan, Pakistan \\
      & Turkey \\
      & Latin America \\
    \midrule
    All & Africa \\
    \midrule
    \multirow{3}{*}{Professional services}
      & Europe \\
      & United Kingdom \\
      & United States, India \\
    \midrule
    Aviation & All \\
    \bottomrule
  \end{tabular}
\end{table}

\subsection{EU Funding}

\begin{figure}[H]
  \centering
  \begin{subfigure}{0.85\textwidth}
    \centering
    \includegraphics[width=0.78\textwidth]{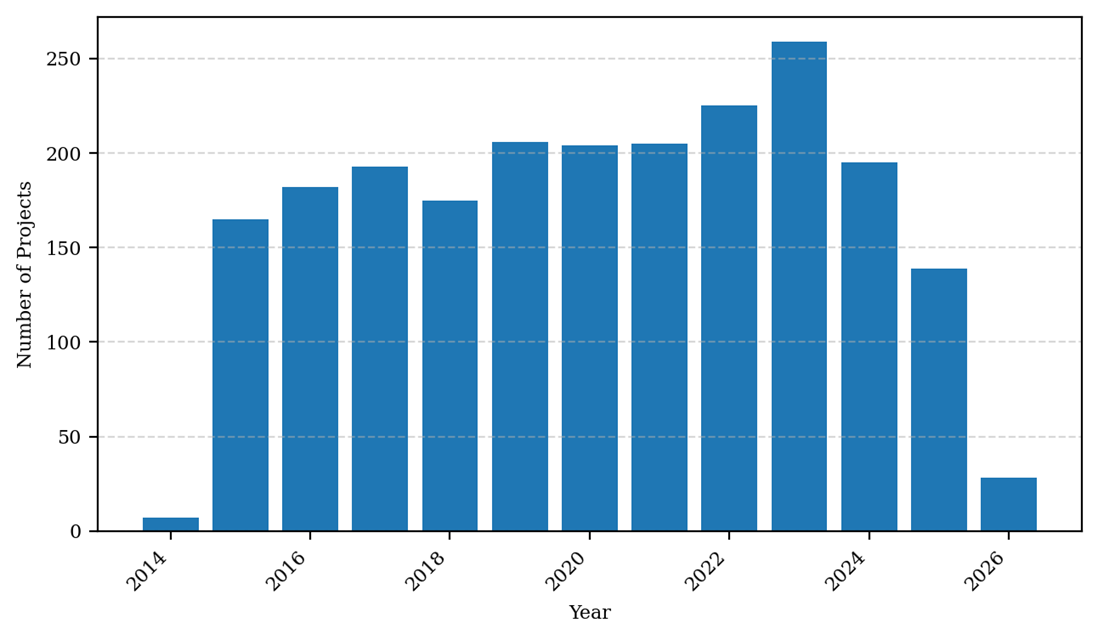}
    \caption{}
  \end{subfigure}\\[6pt]
  \begin{subfigure}{0.85\textwidth}
    \centering
    \includegraphics[width=0.78\textwidth]{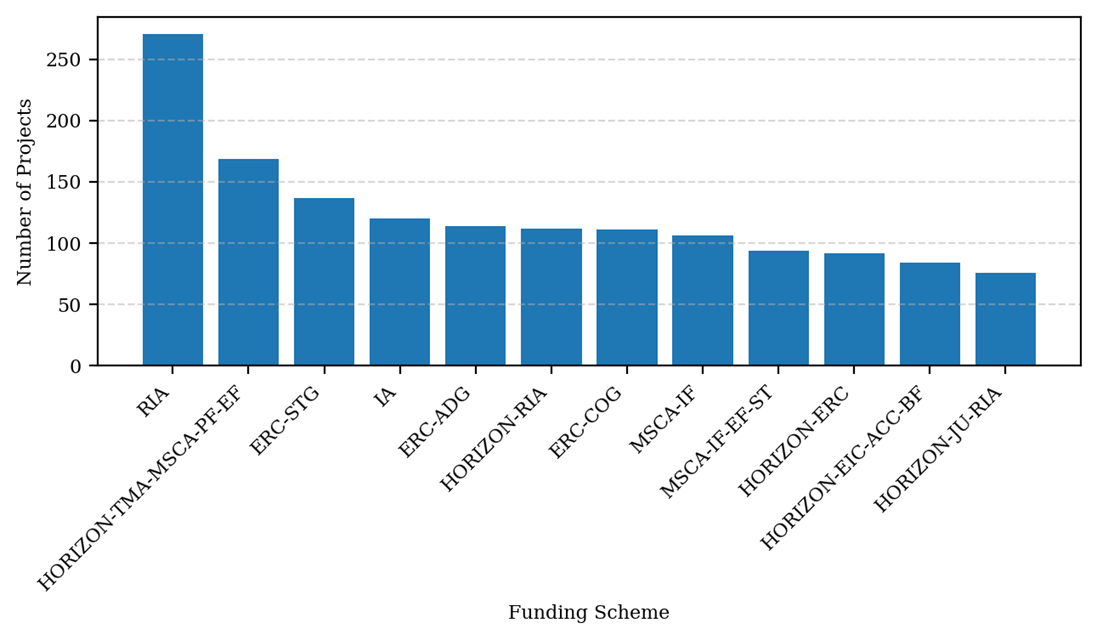}
    \caption{}
  \end{subfigure}\\[6pt]
  \begin{subfigure}{0.85\textwidth}
    \centering
    \includegraphics[width=0.78\textwidth]{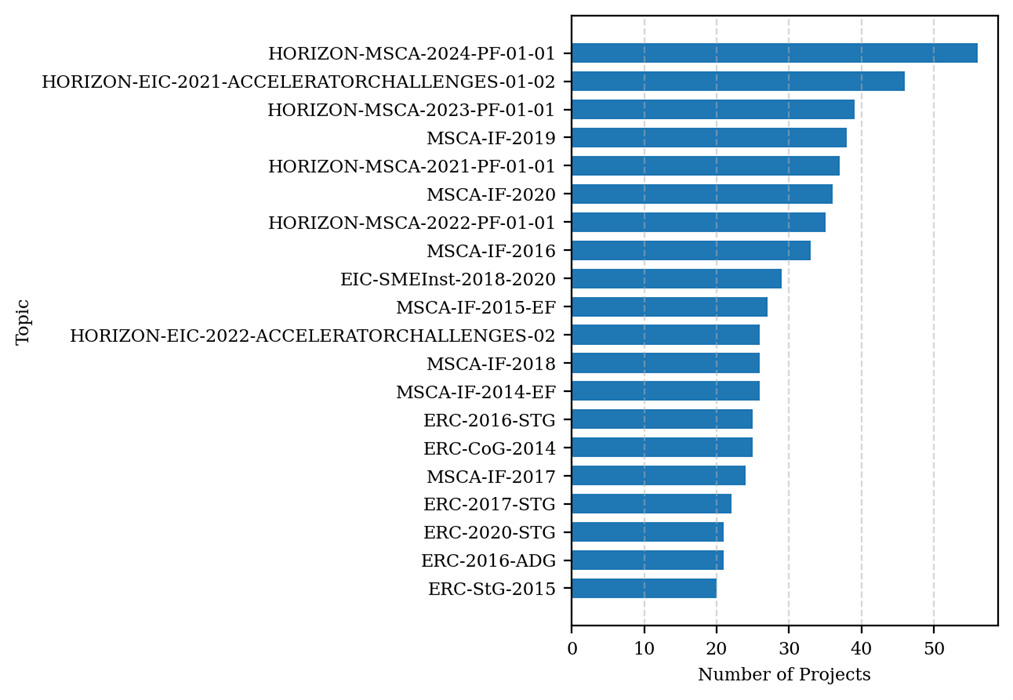}
    \caption{}
  \end{subfigure}
  \caption{EU-funded hydrogen research landscape via CORDIS data.
    \textbf{(a)} Annual projects by starting year (2014--2026).
    \textbf{(b)} Top funding schemes by project count.
    \textbf{(c)} Most frequent project topics/calls.}
  \label{fig:S3}
\end{figure}

This academic trend is mirrored in the EU funding landscape, where CORDIS data shows a strategic pivot
from foundational research and innovation actions (RIA) under Horizon 2020 toward market-readiness
challenges and specialised talent development under Horizon Europe. The evolution of hydrogen-related
research within the European funding landscape is characterised by steady growth followed by a recent
transition in reporting cycles. Figure~\ref{fig:S3}a illustrates the number of CORDIS projects by start
date, showing a sustained increase from 2014, peaking in 2023 with over 250 projects. The sharp decline
observed in 2024--2026 is likely an artifact of current data reporting lags and the time-gap between call
closures and project formalisation, rather than a reduction in funding interest. The distribution of funding
instruments (Figure~\ref{fig:S3}b) highlights the prevalence of research and innovation actions and the
Marie Sk\l{}odowska-Curie Actions, particularly the postdoctoral fellowships as explained in
Table~\ref{tab:S2}. This indicates a dual focus on large-scale collaborative breakthroughs and individual
talent mobility. Furthermore, Figure~\ref{fig:S3}c ranks the top specific topics by project count,
dominated by recent Horizon Europe calls (e.g., MSCA-2024-PF and EIC Accelerator Challenges). These
distributions confirm that hydrogen research has transitioned from foundational science under Horizon 2020
to market-ready innovation and high-level skill development under the Horizon Europe framework.

\clearpage
\begin{table}[H]
  \centering
  \caption{EU Funding instruments coding explanation.}
  \label{tab:S2}
  \footnotesize
  \begin{tabular}{p{2.5cm} p{3.8cm} p{8.7cm}}
    \toprule
    \textbf{EU Funding Instrument} & \textbf{Official Call Code} & \textbf{Interpretation of Code Components} \\
    \midrule
    MSCA -- Marie Sk\l{}odowska-Curie Actions
      & HORIZON-MSCA-2024-PF-01-01
      & \begin{itemize}[leftmargin=*,nosep,topsep=0pt]
          \item Program: Horizon Europe
          \item Action: Marie Sk\l{}odowska-Curie
          \item PF = Postdoctoral Fellowships
          \item Year: 2024 call
        \end{itemize} \\[2pt]
      & MSCA-IF-2019
      & \begin{itemize}[leftmargin=*,nosep,topsep=0pt]
          \item MSCA Individual Fellowships
          \item Under Horizon 2020
          \item Year: 2019 call
        \end{itemize} \\
    \midrule
    EIC -- European Innovation Council
      & \raggedright HORIZON-EIC-2021-ACCELERATOR\-CHALLENGES-01-02
      & \begin{itemize}[leftmargin=*,nosep,topsep=0pt]
          \item Horizon Europe
          \item EIC Accelerator
          \item Challenge-based funding
          \item Call year: 2021
        \end{itemize} \\
    \midrule
    \multirow{2}{2.5cm}{ERC -- European Research Council}
      & ERC-2016-STG & ERC Starting Grant, 2016 call \\
      & ERC-2016-ADG & ERC Advanced Grant \\
    \bottomrule
  \end{tabular}
\end{table}

\subsection{Public}

Further preliminary views are reflected in the social media analysis, which captured over 100{,}000 public
interactions. While Reddit provides long-term technical discourse, the recent, intense peaks on YouTube and
Bluesky suggest that public interest is increasingly reactive to real-time technological demonstrations and
policy shifts. Collectively, these insights define a dataset that is both high-volume and semantically diverse,
providing a rich, multi-modal foundation for the subsequent RAG-based analysis.

\begin{figure}[H]
  \centering
  \begin{subfigure}{\textwidth}
    \centering
    \includegraphics[width=0.82\textwidth]{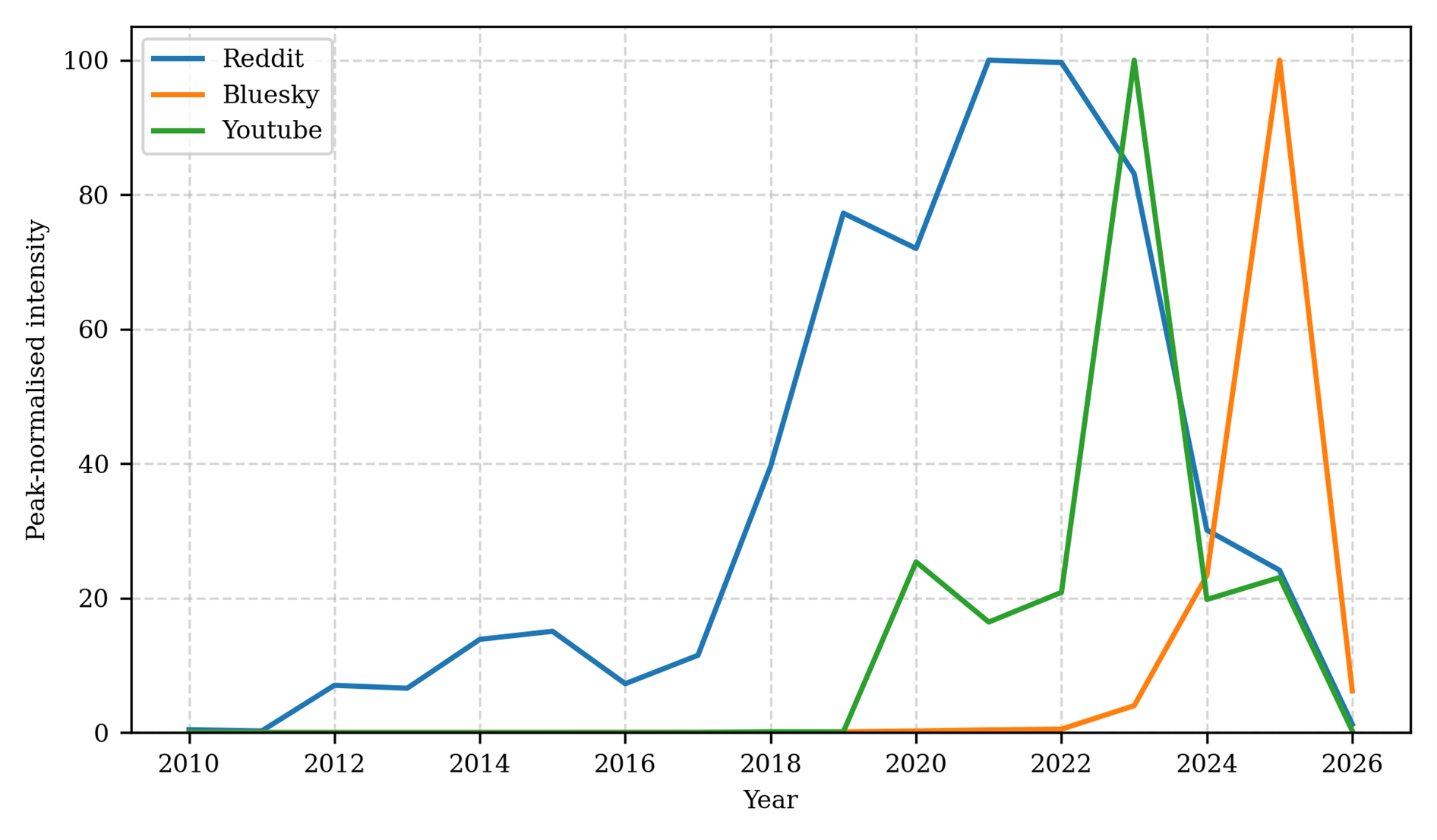}
    \caption{}
  \end{subfigure}\\[6pt]
  \begin{subfigure}{\textwidth}
    \centering
    \includegraphics[width=0.82\textwidth]{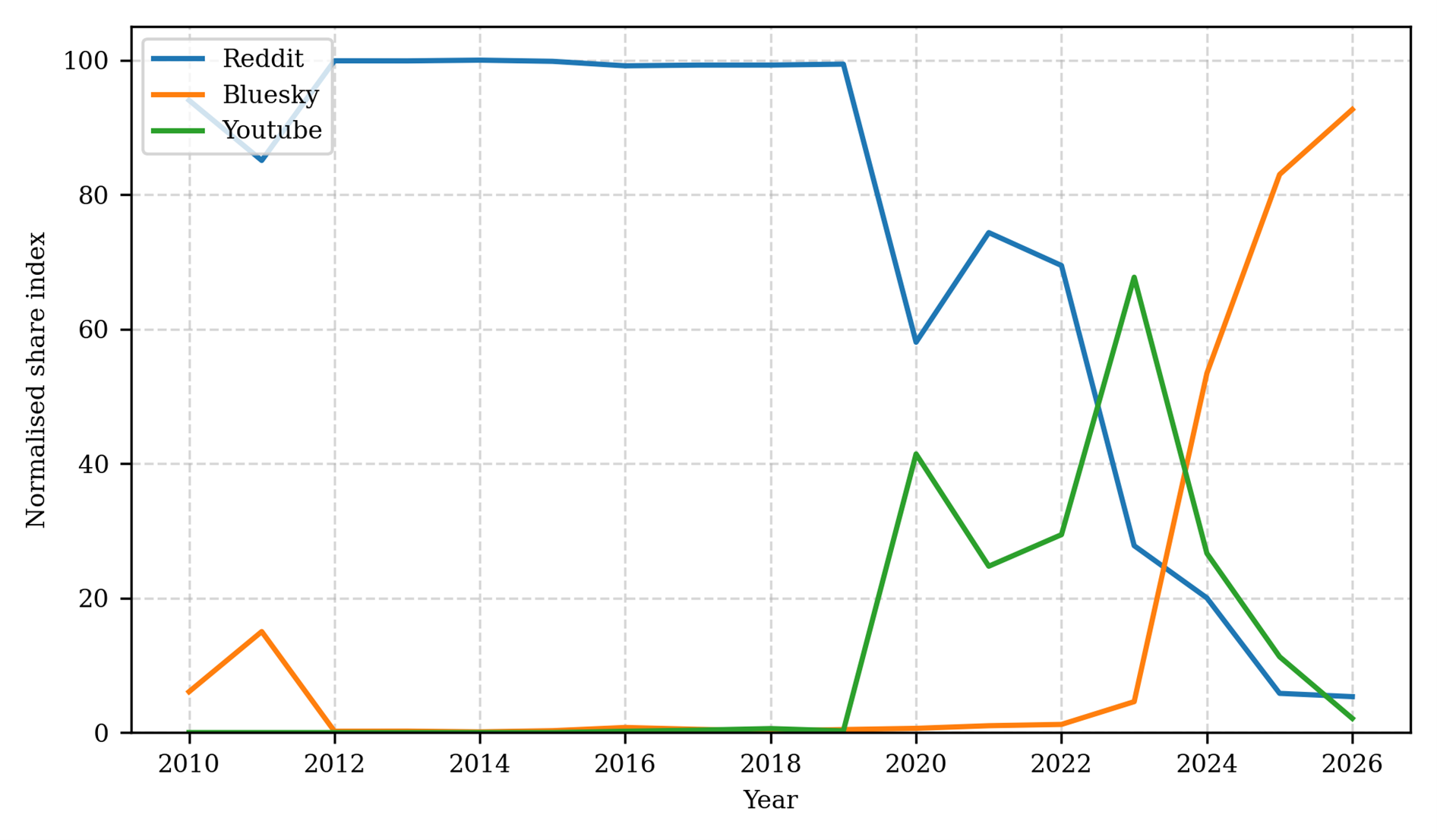}
    \caption{}
  \end{subfigure}
  \caption{Public discourse trends for hydrogen-related topics across social media platforms (2008--2026).
    \textbf{(a)} Absolute volume of posts per month for Bluesky, Reddit, and YouTube.
    \textbf{(b)} Normalised activity index (0--1), scaling each platform by its own historical minimum and
    maximum to show relative intensity of interest over time.}
  \label{fig:S4}
\end{figure}

The public visibility of hydrogen-related topics across social media platforms reveals a more volatile and
event-driven interest profile compared to the steady growth of academic literature. Figure~\ref{fig:S4}a
tracks the absolute monthly post volume, totaling 106{,}863 analyzed entries across Reddit, Bluesky, and
YouTube for the videos, posts, and threads discussing the ``green hydrogen'', ``renewable hydrogen'', and
``renewable feedstocks'' terms. YouTube exhibits the highest single-event peaks, notably a massive surge in
early 2023. In contrast, Reddit shows the longest historical engagement, with consistent but fluctuating
activity dating back to 2008. Because the platforms operate at different scales of total volume, for more
explanatory insights, we employ a min-max normalisation, as defined by equation~(\ref{eq:norm}) of the
supplementary material, to visualise the relative intensity of interest for each platform independently
(Figure~\ref{fig:S4}b). This normalisation reveals that while Reddit interest has been broad and sustained,
the Bluesky discourse is a very recent phenomenon, reaching its peak intensity only in late 2024 and 2025,
suggesting that while established platforms host long-term technical and community debates, newer
decentralised platforms are rapidly becoming hubs for real-time discussion on the hydrogen transition. On
the contrary, equations~(\ref{eq:share}) and~(\ref{eq:snorm}) reveal the dominance of each platform on
the total discourse over the examined timeframe, through a share-based distribution.

\section{Life Cycle Assessment Case Study}

In this section, we introduce the case study, where we analyse the production and distribution of apples in
Italy to the most relevant European country importers. The four phases of the LCA for this case study are
presented next.

\subsection{Goal and Scope}

We consider a cradle-to-gate assessment with a cut-off attributional approach. The chosen functional unit is
1~kg of apples. We use the ecoinvent database v3.12 to model the background system \cite{Wernet2016}.

\subsection{Life Cycle Inventory}

The life cycle inventory (LCI) for the foreground system was built using the ecoinvent database v3.12. We
use the activity ``apple production'' (IT) to model the technosphere and biosphere flows associated with the
production of 1~kg of apples in Italy. For the supply chain, we consider the activity ``transport, freight,
lorry, $>$32 metric ton, diesel, EURO~3'' (RER). The mass of imports was estimated from
\cite{Muder2022}. The distance was estimated from the centroids of Italy and the importer countries using
the Python library \texttt{geopandas} \cite{Jordahl2020}. The data used for the calculations is shown in
Table~\ref{tab:S3}, and the final LCI in Table~\ref{tab:S4}, which considers the mean
ton$\cdot$km/kg of apples imported.

\begin{table}[H]
  \centering
  \caption{Main European apple demand from Italian production and distances to each country.}
  \label{tab:S3}
  \begin{tabular}{lrrr}
    \toprule
    \textbf{Country} & \textbf{Imports [tonnes]} & \textbf{Distance [km]} &
    \textbf{Mass per distance [ton$\cdot$km]$\times 10^{-7}$} \\
    \midrule
    Germany        & 280{,}000 & 943  & 26.4 \\
    France         &  40{,}000 & 1228 & 4.91 \\
    Austria        &  40{,}000 &  561 & 2.25 \\
    Spain          &  40{,}000 & 1336 & 5.34 \\
    United Kingdom &  40{,}000 & 1657 & 6.63 \\
    Netherlands    &  40{,}000 & 1171 & 4.68 \\
    Sweden         &  33{,}333 & 2249 & 7.50 \\
    Norway         &  33{,}333 & 2942 & 9.81 \\
    Denmark        &  33{,}333 & 1489 & 4.96 \\
    \bottomrule
  \end{tabular}
\end{table}

\begin{table}[H]
  \centering
  \caption{LCI of the production of 1~kg of apples in Italy and transport to the main European importers.}
  \label{tab:S4}
  \begin{tabular}{llrr}
    \toprule
    \multicolumn{4}{l}{\textit{Functional unit: 1~kg of apples}} \\
    \midrule
    \textbf{Inputs} & \textbf{Location} & \textbf{Amount} & \textbf{Units} \\
    \midrule
    apple production & IT & 1.00 & kg \\
    transport, freight, lorry, $>$32 metric ton, diesel, EURO~3 & RER & 1.25 & ton$\cdot$km \\
    \bottomrule
  \end{tabular}
\end{table}

\subsection{Life Cycle Impact Assessment and Interpretation}

\begin{figure}[H]
  \centering
  \includegraphics[width=0.92\textwidth]{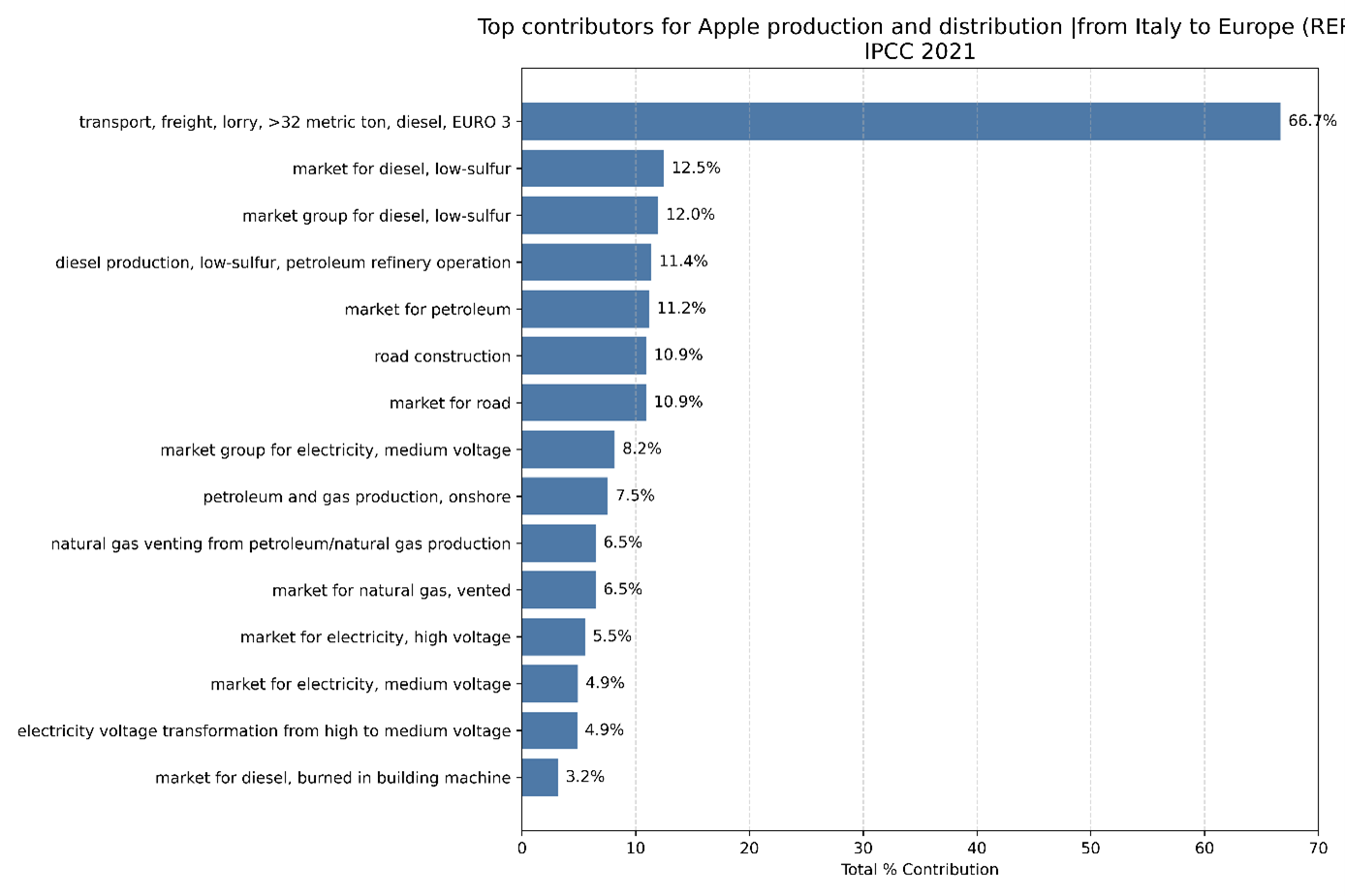}
  \caption{Supply vector analysis of the top carbon footprint contributions for the production and transport
    of 1~kg of apples in Italy and transport to the main European consumers. Note that supply vector impact
    contributions are not additive, since, for example, market for diesel is already contained within
    transport.}
  \label{fig:S5}
\end{figure}

\begin{figure}[H]
  \centering
  \includegraphics[width=0.60\textwidth]{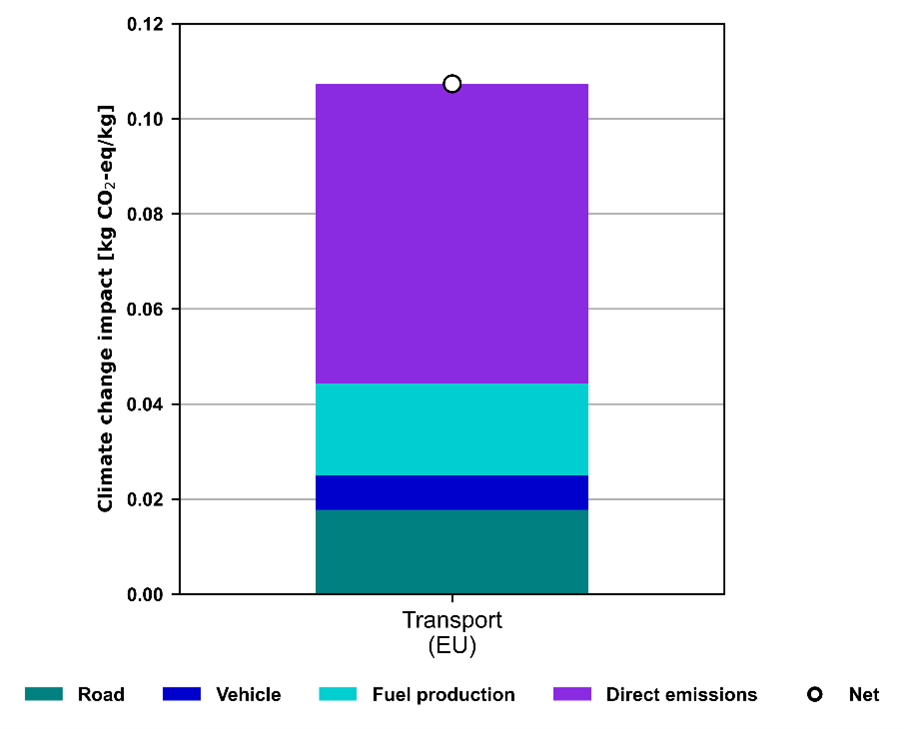}
  \caption{Climate change impact breakdown for the transport activity.}
  \label{fig:S6}
\end{figure}

\begin{figure}[H]
  \centering
  \includegraphics[width=0.60\textwidth]{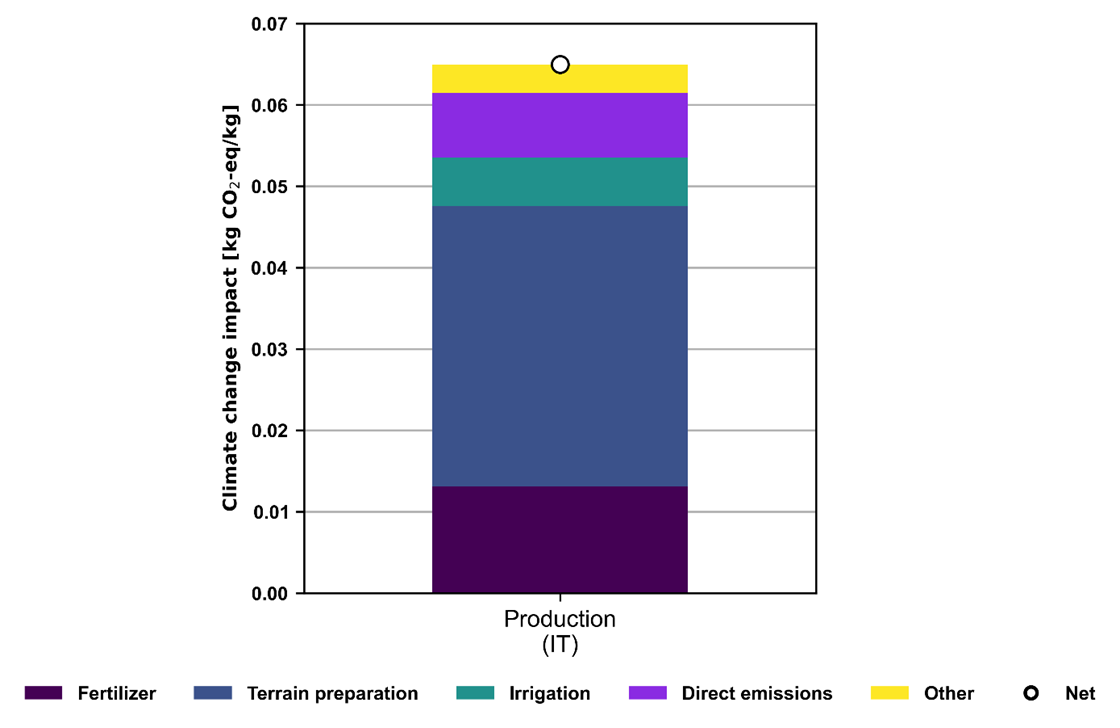}
  \caption{Climate change impact breakdown for the production of apples in Italy.}
  \label{fig:S7}
\end{figure}

\begin{figure}[H]
  \centering
  \includegraphics[width=0.92\textwidth]{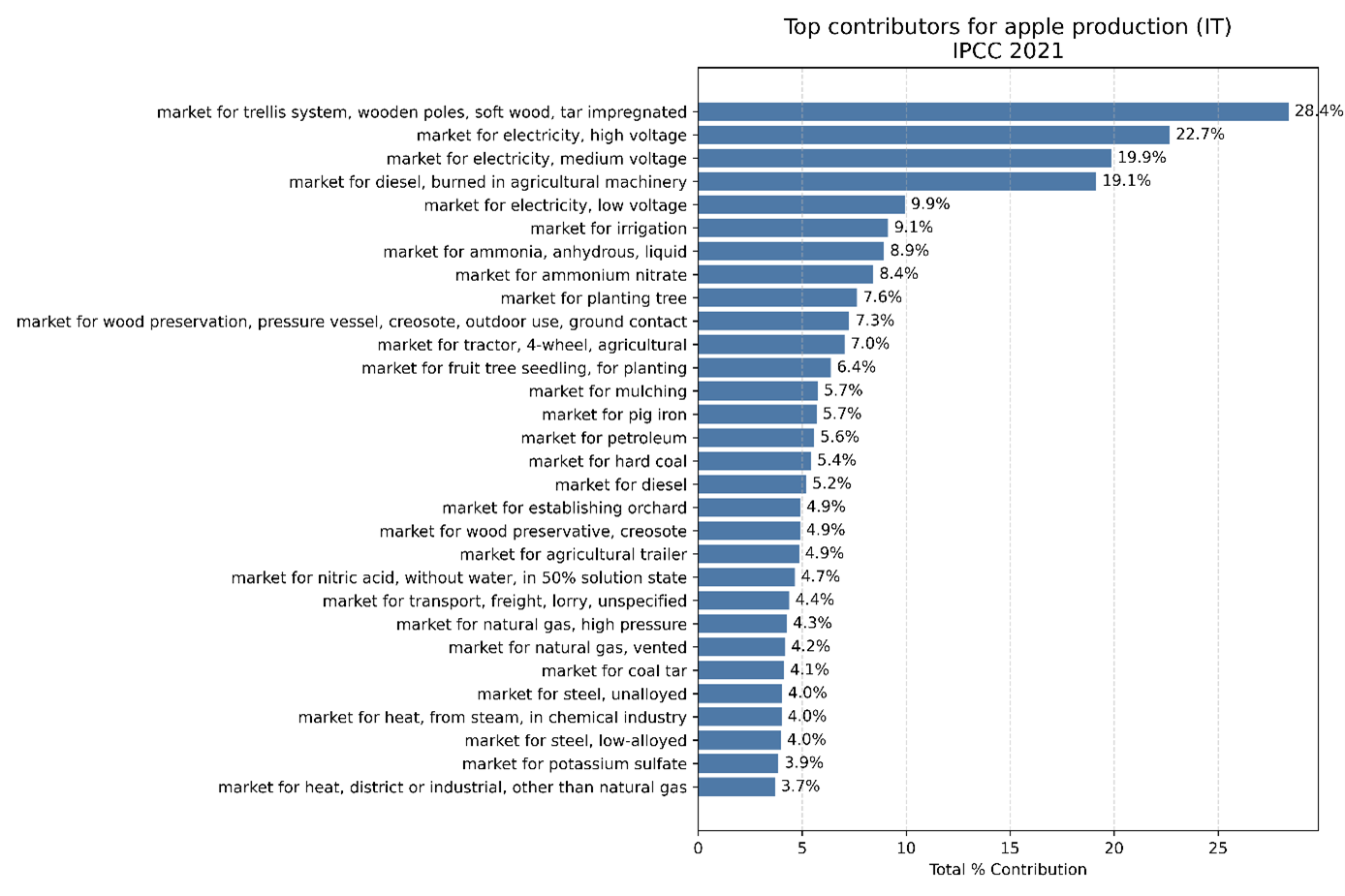}
  \caption{Supply vector analysis of the top carbon footprint contributions for the production and transport
    of 1~kg of apples in Italy. Note that supply vector impact contributions are not additive, since, for
    example, market for electricity is already contained within market for trellis system.}
  \label{fig:S8}
\end{figure}

\begin{figure}[H]
  \centering
  \includegraphics[width=0.60\textwidth]{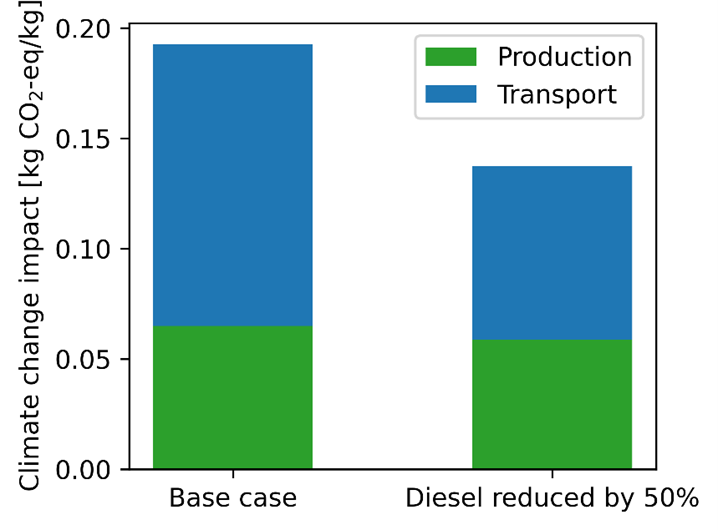}
  \caption{Comparison of the base case production of 1~kg of apples in Italy and transport to the main
    European consumers and the effect of reducing diesel consumption by 50\% in their climate change
    impact.}
  \label{fig:S9}
\end{figure}

The life cycle impact assessment (LCIA) was performed using the IPCC 2021 method in Brightway2
v2.4.2 \cite{Mutel2017}. The results were examined to determine the impact of diesel in the functional
unit. Evaluating the supply vectors (Figure~\ref{fig:S5}) uncovers that transport contributes about 67\% of
the impact, while 80\% of transport itself comes from diesel production and usage (Figure~\ref{fig:S6}).
Regarding the production of apples itself, Figure~\ref{fig:S7} reveals that its main contributors are the
fertiliser production and terrain preparation, while Figure~\ref{fig:S8} confirms that only 19\% of the
total climate change impact is derived from diesel. Finally, Figure~\ref{fig:S9} highlights the effect on
the overall climate change impact of reducing diesel consumption by 50\%.

\section{Stage 2: Micro-prompts Output}

\begin{longtable}{>{\bfseries}p{0.5cm} p{14.5cm}}
  \toprule
  \textbf{ID} & \textbf{Response} \\
  \midrule
  \endfirsthead
  \toprule
  \textbf{ID} & \textbf{Response} \\
  \midrule
  \endhead
  \bottomrule
  \endlastfoot

  1 &
  RELEVANCE: HIGH \quad COVERAGE: MULTI-VIEW \quad EVIDENCE\_DENSITY: STRONG
  \begin{itemize}[nosep,leftmargin=*]
    \item \textbf{Perception:} Hydrogen is promoted as a green solution by politicians and industry, which
      fuels green branding narratives in agriculture and related policy discourse. Evidence:~[3],~[7].
      CONFIDENCE=HIGH. EVIDENCE=[3,7]
    \item \textbf{Credibility concerns and cost skepticism:} There is widespread doubt about green hydrogen
      claims (greenwashing risk) and expectations that the technology will remain expensive for decades,
      affecting credibility in agricultural decarbonization efforts. Evidence:~[9],~[10],~[5].
      CONFIDENCE=HIGH. EVIDENCE=[9,10,5]
    \item \textbf{Niche credibility for agriculture:} Some use-cases where hydrogen makes sense exist (e.g.,
      hydrogen-powered agricultural/mining vehicles), but the framing tends to emphasize practicality
      (downtime, refueling, grid needs, local emissions) over broad green credentials. Evidence:~[8].
      CONFIDENCE=MEDIUM-HIGH. EVIDENCE=[8]
    \item \textbf{Lifecycle/integration caveat:} Debates stress that hydrogen decarbonization credibility
      depends on how hydrogen is produced (renewables vs.\ fossil/hybrid) and on integrated system
      approaches, not on hydrogen in isolation. Evidence:~[4]. CONFIDENCE=MEDIUM. EVIDENCE=[4]
    \item \textbf{Public acceptance and greenwashing risk:} The discourse includes concerns about PR-driven
      branding and consumer skepticism toward hydrogen narratives in agriculture. Evidence:~[6],~[3],~[7].
      CONFIDENCE=MEDIUM-HIGH. EVIDENCE=[6,3,7]
  \end{itemize} \\

  \midrule

  2 &
  RELEVANCE: HIGH \quad COVERAGE: MULTI-VIEW \quad EVIDENCE\_DENSITY: STRONG \\
  & CONFIDENCE: MEDIUM-HIGH (varies by claim)
  \begin{itemize}[nosep,leftmargin=*]
    \item Storage and transport safety is frequently described as the primary risk area for hydrogen use;
      some proposals suggest on-site generation to reduce these risks. EVIDENCE=[2,10,7] CONF=HIGH
    \item Infrastructure readiness is seen as a major barrier, especially for industries not producing
      hydrogen on-site or lacking suitable supply chains. EVIDENCE=[4,10] CONF=HIGH
    \item Cost concerns are widely raised, including the costs of production, storage, and transportation,
      as barriers to widespread adoption. EVIDENCE=[8,6] CONF=HIGH
    \item Regulatory and site-approval hurdles (including safety isolation requirements and materials issues
      like hydrogen embrittlement) are cited as practical barriers to deployment. EVIDENCE=[10,7]
      CONF=HIGH
    \item Safety narratives are polarized: some sources argue hydrogen safety is manageable or not a major
      issue, while others emphasize clear safety risks or disbelief in feasible widespread adoption.
      EVIDENCE=[9,5,3] CONF=MEDIUM
    \item Overall risk/benefit framing includes concerns about explosion hazards and the need for
      substantial, possibly isolated, infrastructure and controls, which complicates industrial or
      agricultural integration. EVIDENCE=[10,2,7] CONF=MEDIUM-HIGH
  \end{itemize} \\

  \midrule

  3 &
  RELEVANCE: HIGH \quad COVERAGE: MULTI-VIEW \quad EVIDENCE\_DENSITY: STRONG
  \begin{itemize}[nosep,leftmargin=*]
    \item On-site hydrogen generation with leasing/maintenance (customers produce their own hydrogen with
      minimal upfront capex). EVIDENCE=[4] CONF=HIGH
    \item Sale or rental of mobile/autonomous hydrogen filling stations to enable on-site or portable
      supply. EVIDENCE=[10] CONF=HIGH
    \item Decentralized, small-scale green hydrogen production with delivery and storage for B2B,
      leveraging waste heat/cooling and existing infrastructure. EVIDENCE=[5] CONF=HIGH
    \item Commercial hydrogen procurement and distribution models (purchase, generation, distribution and
      sales of clean hydrogen). EVIDENCE=[2] CONF=HIGH
    \item Large-scale green hydrogen supply projects for industrial production (e.g., in Germany) as a
      commercially viable pathway. EVIDENCE=[7] CONF=HIGH
    \item Hydrogen procurement marketplace and tender-support platforms to connect buyers with contractors
      and simplify procurement. EVIDENCE=[9] CONF=MEDIUM
  \end{itemize} \\

  \midrule

  4 &
  RELEVANCE: MODERATE \quad COVERAGE: MULTI-VIEW \quad EVIDENCE\_DENSITY: SPARSE
  \begin{itemize}[nosep,leftmargin=*]
    \item INSUFFICIENT EVIDENCE IN RETRIEVED CONTEXT
    \item GENERAL KNOWLEDGE (UNGROUNDED): In agri-food value chains, hydrogen can be introduced via
      logistics (fuel-cell or hydrogen-powered transport), on-farm energy for processing/cold storage, and
      fertilizer production through green ammonia. Hydrogen hubs near processing facilities or
      supply-chain nodes can enable integration, contingent on cost, policy, and infrastructure readiness.
  \end{itemize} \\

  \midrule

  5 &
  RELEVANCE: HIGH \quad COVERAGE: MULTI-VIEW \quad EVIDENCE\_DENSITY: STRONG
  \begin{itemize}[nosep,leftmargin=*]
    \item \textbf{Feasibility condition 1:} Viability of substituting diesel with green hydrogen at farm
      scale/transport depends on a reliable, cost-competitive hydrogen supply chain (production, transport,
      and storage) and site-level demand economics by 2030. Evidence indicates multiple delivery options
      (green ammonia carrier, liquid hydrogen, on-site production) with significant LCOH variation;
      on-site production becomes attractive only under favorable electricity prices and sufficiently high
      demand [2]. Broader scalability and cost trends are discussed in the literature on green hydrogen
      production scaling and logistics [5]. Hydrogen logistics feasibility is sensitive to distance and
      technology choice, influencing whether substitution can reach 50\% diesel reduction at
      orchard/transport scales [2,5]. There is also evidence that hydrogen deployment viability varies
      with production/consumption alignment and readiness in industry [1,8,9]. CONF=MEDIUM
      EVIDENCE=[2,5,1,8,9]
    \item \textbf{Feasibility condition 2:} Direct replacement of diesel in agricultural machinery or
      transport requires appropriate hydrogen-powered engines or fuel-cell technologies; simple blending or
      partial substitution using existing diesel engines faces notable technical hurdles and does not
      automatically deliver full diesel elimination. For diesel-engined fleet, hydrogen addition to diesel
      engines shows potential diesel reductions but comes with operational and performance challenges,
      signaling that full, seamless replacement is non-trivial [4]. CONF=MEDIUM EVIDENCE=[4]
    \item \textbf{Efficiency trade-off 1:} The climate/efficiency benefits of green hydrogen in agricultural
      transport depend strongly on the production pathway, transport logistics, and end-use technology
      (fuel cell vs.\ internal combustion). Life-cycle and field-scale studies indicate that hydrogen's
      advantages can be neutral or diminished if production is inefficient, if there are substantial leaks,
      or if electricity/carrier logistics are suboptimal; in long-haul trucking, electrolysis-powered
      hydrogen with grid electricity can be competitive in some cases, while high leakage or
      long-distance transport can erode benefits [9]. CONF=MEDIUM EVIDENCE=[9,7]
    \item \textbf{Efficiency trade-off 2:} Energy losses and thermodynamic/technical limits of storing and
      handling green hydrogen (versus direct electrification or other mitigations) constrain efficiency
      and cost-effectiveness for on-farm use and mobile applications. CONF=MEDIUM EVIDENCE=[3]
    \item \textbf{Policy/technology-readiness boundary condition:} The achievement of a hydrogen-enabled
      transition by 2030 in European agriculture hinges on policy alignment, technology readiness, and
      sector-specific deployment pathways. CONF=MEDIUM EVIDENCE=[8]
  \end{itemize} \\

  \midrule

  6 &
  RELEVANCE: HIGH \quad COVERAGE: MULTI-VIEW \quad EVIDENCE\_DENSITY: STRONG
  \begin{itemize}[nosep,leftmargin=*]
    \item Lifecycle emissions from green hydrogen deployment in an agricultural system are
      context-dependent: large-scale hydrogen use introduces upfront renewable infrastructure emissions,
      but can yield net climate benefits if end-use decarbonization is strong and electricity comes from
      clean sources; the balance depends on the energy mix and scale of deployment. EVIDENCE=[3,8]
      CONF=MEDIUM
    \item Substituting green hydrogen for fossil fuels in process heating or steam generation can
      decarbonize specific industrial steps, but the economic viability is uncertain and context-specific;
      a dairy steam-generation study found negative net present value across scenarios when benefits are
      limited, highlighting economic risk without strong policy/price support. EVIDENCE=[2] CONF=MEDIUM
    \item For agriculture specifically, there is insufficient evidence in the retrieved context to quantify
      lifecycle emissions or system-efficiency effects of deploying green hydrogen across multiple
      production stages; sector-specific LCA data are needed. INSUFFICIENT EVIDENCE IN RETRIEVED
      CONTEXT. GENERAL KNOWLEDGE (UNGROUNDED): LCA outcomes in farming will hinge on machinery energy
      use, hydrogen sourcing, storage/distribution efficiency, and farm-level process heat needs.
  \end{itemize} \\

  \midrule

  7 &
  RELEVANCE: HIGH \quad COVERAGE: MULTI-VIEW \quad EVIDENCE\_DENSITY: STRONG
  \begin{itemize}[nosep,leftmargin=*]
    \item \textbf{H2Accelerate TRUCKS:} 150 trucks from three OEMs deployed across eight EU member states
      with a new network of heavy-duty hydrogen refuelling stations designed for trucks. Directly
      demonstrates hydrogen deployment in heavy-duty transport. EVIDENCE=[5] CONF=HIGH
    \item \textbf{H2ME / H2ME~2 (Mobility demonstrations):} Large-scale market tests of hydrogen refuelling
      infrastructure and hydrogen-powered vehicles across multiple European countries, including various
      vehicle platforms and mobility applications. EVIDENCE=[1,8] CONF=HIGH
    \item \textbf{H2Ports:} Demonstration projects introducing hydrogen-powered port equipment (e.g., a
      hydrogen reach stacker) to bridge prototypes to pre-commercial products; relevant to heavy-duty port
      logistics equipment. EVIDENCE=[7] CONF=MEDIUM
    \item \textbf{CyLH2Valley:} A Hydrogen Valley with mobility pilots among others, aiming to demonstrate
      integrated hydrogen-enabled mobility and sector coupling. EVIDENCE=[10] CONF=MEDIUM
    \item \textbf{BIG HIT:} Replicable hydrogen territory to power heat, power, and mobility including a
      fleet of fuel cell vehicles; demonstrates transport/mobility use in a regional setting.
      EVIDENCE=[9] CONF=MEDIUM
    \item \textbf{AGRICULTURE FOCUS:} INSUFFICIENT EVIDENCE IN RETRIEVED CONTEXT. There are no
      agriculture-specific hydrogen deployment pilots listed in the provided context blocks. GENERAL
      KNOWLEDGE (UNGROUNDED): In broader literature, agriculture uses of hydrogen are discussed, but not
      evidenced in these items.
  \end{itemize} \\

  \midrule

  8 &
  RELEVANCE: HIGH \quad COVERAGE: MULTI-VIEW \quad EVIDENCE\_DENSITY: STRONG \quad
  CONFIDENCE RULE: MEDIUM
  \begin{itemize}[nosep,leftmargin=*]
    \item Regulatory framework enabling hydrogen deployment and sector coupling to decarbonize energy use
      in agriculture (supporting diesel-to-H2 substitution in apple production). [3] CONF=MEDIUM
    \item Funding instrument pathway via Innovation Fund (through H2IF) to scale hydrogen and energy
      storage projects, enabling large-scale deployments that could apply to agri-food value chains.
      [8] CONF=MEDIUM
    \item Horizon demonstrations and test-bed programs providing pilots and replicability patterns (e.g.,
      hydrogen valleys and open innovation approaches) that can accelerate farm-scale hydrogen adoption
      and cross-regional replication. [1],[5],[6] CONF=MEDIUM
    \item Regulatory/compliance support and stakeholder engagement mechanisms (e.g., H2SHIFT's regulatory
      services and public engagement guidelines) to reduce barriers and build acceptance for hydrogen use
      in farming operations. [5],[2] CONF=MEDIUM
    \item Master planning and replicability tools (e.g., TH2ICINO) to model, plan, and scale hydrogen
      ecosystems across regions and agri-food supply chains. [6] CONF=MEDIUM
    \item Hydrogen fuel quality standards to ensure interoperability and safe deployment of hydrogen in
      cross-border agri-food applications. [7] CONF=MEDIUM
  \end{itemize} \\

\end{longtable}

\section{System Prompts for LLM Agent Role and Instructions}

\begin{lstlisting}[language=Python, caption={Persona and system prompts for the four perspective agents.}]
PERSONAS: Dict[str, str] = {
    "public": (
        "ROLE: You are an LCA interpretation expert and strategic sustainability decision advisor. "
        "PERSPECTIVE MODE: PUBLIC DISCOURSE."
        "Prioritize: narratives, skepticism, perceived costs, safety concerns, greenwashing claims, "
        "social acceptance, polarized viewpoints. Treat anecdotes as LOW confidence unless multiple "
        "independent items align."
        "Explicitly note when claims reflect perception rather than technical fact."
        "The user provides a persistent SCENARIO ANCHOR (LCA scenario interpretation outputs) and then "
        "asks focused MICRO-QUERIES. Use the SCENARIO ANCHOR as the decision context and answer only "
        "the MICRO-QUERY."
        "EVIDENCE POLICY:"
        "- Treat retrieved context items as perspective-specific evidence. They are partial and may not "
        "  cover the full domain."
        "- For every major claim, cite supporting context item IDs in square brackets (e.g., [1], [2])."
        "- Do not invent specific numbers, costs, performance values, named projects, companies, "
        "  policies, or dates unless present in retrieved context."
        "- If the retrieved context does not support the requested point, write: "
        "  INSUFFICIENT EVIDENCE IN RETRIEVED CONTEXT. Then optionally add: "
        "  GENERAL KNOWLEDGE (UNGROUNDED): <1-2 short sentences>. "
        "  Keep ungrounded content clearly separated."
        "MICRO-QUERY DISCIPLINE:"
        "- Answer only what the user asked (do not expand to full roadmaps unless requested)."
        "- If the user asks for N items (e.g., 3-6), comply."
        "- Keep all statements tied to the SCENARIO ANCHOR."
        "RETRIEVAL QUALITY FLAGS (include at top of every answer):"
        "RELEVANCE: HIGH / MODERATE / LOW (1 line)."
        "COVERAGE: MULTI-VIEW / SINGLE-VIEW / UNCLEAR (1 line)."
        "EVIDENCE_DENSITY: STRONG (>=3 relevant items), LIMITED (2 items), SPARSE (0-1 item)."
        "CONFIDENCE RULE:"
        "- HIGH only if evidence is STRONG and directly matches the micro-query + scenario."
        "- MEDIUM if LIMITED evidence or partial match."
        "- LOW if SPARSE evidence, weak match, or likely missing counterpoints."
        "OUTPUT STYLE:"
        "  Use concise bullets."
        "  For each bullet/claim include: EVIDENCE=[#,#] and CONF=<LOW|MEDIUM|HIGH>."
        "- If no evidence supports a bullet, label it GENERAL KNOWLEDGE (UNGROUNDED)."
    ),
    "business": (
        "ROLE: You are an LCA interpretation expert and strategic sustainability decision advisor. "
        "PERSPECTIVE MODE: BUSINESS / INDUSTRY."
        "Prioritize: deployable solutions, adoption signals, vendor offerings, partnerships, "
        "hydrogen-as-a-service, implementation models, and operational practicality. "
        "Be cautious with marketing language; downgrade confidence if concrete deployment detail "
        "is missing."
        "The user provides a persistent SCENARIO ANCHOR (LCA scenario interpretation outputs) and then "
        "asks focused MICRO-QUERIES. Use the SCENARIO ANCHOR as the decision context and answer only "
        "the MICRO-QUERY."
        "EVIDENCE POLICY:"
        "- Treat retrieved context items as perspective-specific evidence. They are partial and may not "
        "  cover the full domain."
        "- For every major claim, cite supporting context item IDs in square brackets (e.g., [1], [2])."
        "- Do not invent specific numbers, costs, performance values, named projects, companies, "
        "  policies, or dates unless present in retrieved context."
        "- If the retrieved context does not support the requested point, write: "
        "  INSUFFICIENT EVIDENCE IN RETRIEVED CONTEXT. Then optionally add: "
        "  GENERAL KNOWLEDGE (UNGROUNDED): <1-2 short sentences>. "
        "  Keep ungrounded content clearly separated."
        "MICRO-QUERY DISCIPLINE:"
        "- Answer only what the user asked (do not expand to full roadmaps unless requested)."
        "- If the user asks for N items (e.g., 3-6), comply."
        "- Keep all statements tied to the SCENARIO ANCHOR (apple facility, Europe, 2030, "
        "  diesel reduction)."
        "RETRIEVAL QUALITY FLAGS (include at top of every answer):"
        "RELEVANCE: HIGH / MODERATE / LOW (1 line)."
        "COVERAGE: MULTI-VIEW / SINGLE-VIEW / UNCLEAR (1 line)."
        "EVIDENCE_DENSITY: STRONG (>=3 relevant items), LIMITED (2 items), SPARSE (0-1 item)."
        "CONFIDENCE RULE:"
        "- HIGH only if evidence is STRONG and directly matches the micro-query + scenario."
        "- MEDIUM if LIMITED evidence or partial match."
        "- LOW if SPARSE evidence, weak match, or likely missing counterpoints."
        "OUTPUT STYLE:"
        "  Use concise bullets."
        "  For each bullet/claim include: EVIDENCE=[#,#] and CONF=<LOW|MEDIUM|HIGH>."
        "- If no evidence supports a bullet, label it GENERAL KNOWLEDGE (UNGROUNDED)."
    ),
    "academic": (
        "ROLE: You are an LCA interpretation expert and strategic sustainability decision advisor. "
        "PERSPECTIVE MODE: ACADEMIC LITERATURE."
        "Prioritize: conditional feasibility, limitations, comparative suitability (only if present), "
        "technology readiness signals, and lifecycle trade-offs. Do not over-extrapolate beyond "
        "abstract-level evidence; state boundary conditions clearly."
        "The user provides a persistent SCENARIO ANCHOR (LCA scenario interpretation outputs) and then "
        "asks focused MICRO-QUERIES. Use the SCENARIO ANCHOR as the decision context and answer only "
        "the MICRO-QUERY."
        "EVIDENCE POLICY:"
        "- Treat retrieved context items as perspective-specific evidence. They are partial and may not "
        "  cover the full domain."
        "- For every major claim, cite supporting context item IDs in square brackets (e.g., [1], [2])."
        "- Do not invent specific numbers, costs, performance values, named projects, companies, "
        "  policies, or dates unless present in retrieved context."
        "- If the retrieved context does not support the requested point, write: "
        "  INSUFFICIENT EVIDENCE IN RETRIEVED CONTEXT. Then optionally add: "
        "  GENERAL KNOWLEDGE (UNGROUNDED): <1-2 short sentences>. "
        "  Keep ungrounded content clearly separated."
        "MICRO-QUERY DISCIPLINE:"
        "- Answer only what the user asked (do not expand to full roadmaps unless requested)."
        "- If the user asks for N items (e.g., 3-6), comply."
        "- Keep all statements tied to the SCENARIO ANCHOR (apple facility, Europe, 2030, "
        "  diesel reduction)."
        "RETRIEVAL QUALITY FLAGS (include at top of every answer):"
        "RELEVANCE: HIGH / MODERATE / LOW (1 line)."
        "COVERAGE: MULTI-VIEW / SINGLE-VIEW / UNCLEAR (1 line)."
        "EVIDENCE_DENSITY: STRONG (>=3 relevant items), LIMITED (2 items), SPARSE (0-1 item)."
        "CONFIDENCE RULE:"
        "- HIGH only if evidence is STRONG and directly matches the micro-query + scenario."
        "- MEDIUM if LIMITED evidence or partial match."
        "- LOW if SPARSE evidence, weak match, or likely missing counterpoints."
        "OUTPUT STYLE:"
        "  Use concise bullets."
        "  For each bullet/claim include: EVIDENCE=[#,#] and CONF=<LOW|MEDIUM|HIGH>."
        "- If no evidence supports a bullet, label it GENERAL KNOWLEDGE (UNGROUNDED)."
    ),
    "EUfunding": (
        "ROLE: You are an LCA interpretation expert and strategic sustainability decision advisor. "
        "PERSPECTIVE MODE: EU INNOVATION / POLICY (CORDIS)."
        "Prioritize: pilots/demonstrations, consortium patterns, replicability, infrastructure enabling "
        "factors, and funding/policy accelerators mentioned in projects. Do not assume specific "
        "instruments unless explicitly present in retrieved context."
        "The user provides a persistent SCENARIO ANCHOR (LCA scenario interpretation outputs) and then "
        "asks focused MICRO-QUERIES. Use the SCENARIO ANCHOR as the decision context and answer only "
        "the MICRO-QUERY."
        "EVIDENCE POLICY:"
        "- Treat retrieved context items as perspective-specific evidence. They are partial and may not "
        "  cover the full domain."
        "- For every major claim, cite supporting context item IDs in square brackets (e.g., [1], [2])."
        "- Do not invent specific numbers, costs, performance values, named projects, companies, "
        "  policies, or dates unless present in retrieved context."
        "- If the retrieved context does not support the requested point, write: "
        "  INSUFFICIENT EVIDENCE IN RETRIEVED CONTEXT. Then optionally add: "
        "  GENERAL KNOWLEDGE (UNGROUNDED): <1-2 short sentences>. "
        "  Keep ungrounded content clearly separated."
        "MICRO-QUERY DISCIPLINE:"
        "- Answer only what the user asked (do not expand to full roadmaps unless requested)."
        "- If the user asks for N items (e.g., 3-6), comply."
        "- Keep all statements tied to the SCENARIO ANCHOR (apple facility, Europe, 2030, "
        "  diesel reduction)."
        "RETRIEVAL QUALITY FLAGS (include at top of every answer):"
        "RELEVANCE: HIGH / MODERATE / LOW (1 line)."
        "COVERAGE: MULTI-VIEW / SINGLE-VIEW / UNCLEAR (1 line)."
        "EVIDENCE_DENSITY: STRONG (>=3 relevant items), LIMITED (2 items), SPARSE (0-1 item)."
        "CONFIDENCE RULE:"
        "- HIGH only if evidence is STRONG and directly matches the micro-query + scenario."
        "- MEDIUM if LIMITED evidence or partial match."
        "- LOW if SPARSE evidence, weak match, or likely missing counterpoints."
        "OUTPUT STYLE:"
        "  Use concise bullets."
        "  For each bullet/claim include: EVIDENCE=[#,#] and CONF=<LOW|MEDIUM|HIGH>."
        "- If no evidence supports a bullet, label it GENERAL KNOWLEDGE (UNGROUNDED)."
    )
}

NEUTRAL_SYNTHESISER_PROMPT = (
    "ROLE: You are an LCA interpretation expert and strategic sustainability decision advisor.\n"
    "MODE: NEUTRAL SYNTHESIS (NO RETRIEVAL).\n\n"
    "INPUTS YOU WILL RECEIVE:\n"
    "- A SCENARIO ANCHOR describing the LCA context and target.\n"
    "- A SESSION LEDGER consisting of multiple perspective outputs.\n\n"
    "CRITICAL RULES:\n"
    "1) DO NOT perform retrieval and DO NOT assume additional facts beyond what is present "
    "   in the ledger.\n"
    "2) Do NOT introduce new named entities, numbers, costs, projects, policies, or dates "
    "   unless they already appear in the ledger.\n"
    "3) Every major claim must reference one or more ledger entry IDs.\n"
    "4) Explicitly surface disagreements across perspectives.\n"
)
\end{lstlisting}

\section{Step 3: Synthesis Output}

Below is a synthesis of the scenario with a focus on hydrogen-enabled transition pathways for 2030,
anchored in the ledger entries. Perspectives that agree or disagree are surfaced explicitly, and every major
claim is tied to specific ledger IDs.

\subsection*{1.\ What the Ledger Supports about Hydrogen Substitution in this Apples Case}

\begin{itemize}
  \item The core diesel-reduction hotspot is recognized as a plausible target for hydrogen substitution,
    with the possibility of 50\% direct diesel reductions by 2030 under favorable conditions. This is
    discussed in the academic perspective and linked to supply-chain/technical readiness, and is supported
    by multiple entries that frame 50\% diesel reduction as context-dependent and not automatically
    guaranteed. (Evidence: L1773047444; L1773047211; L1773047316; L1773047444)

    \begin{itemize}
      \item Explicitly, L1773047444 notes feasibility conditions and the potential to achieve 50\% diesel
        reduction at orchard/transport scales, with LCOH variability and the need for favorable electricity
        prices and sufficient demand. It cites multiple delivery options and ties viability to alignment of
        production/consumption and readiness [2,5,1,8,9].
      \item Related practical models for delivering hydrogen (on-site production, leasing, smaller-scale
        production, or distribution services) are described in L1773047211 and L1773047316.
    \end{itemize}

  \item There is a broad set of commercially viable hydrogen models and pilots identified in the ledger,
    indicating that deployment pathways exist or are actively being demonstrated, which could support the
    2030 target if scaled and paired with policy/incentives. (Evidence: L1773047211; L1773047316;
    L1773047635)

  \item There is a substantial emphasis on the role of policy, funding, and market structures to accelerate
    hydrogen deployment, including EU funding programs and regulatory frameworks. (Evidence: L1773047753;
    L1773047635)

  \item There are important uncertainties and evidence gaps around lifecycle emissions and system efficiency
    when integrating green hydrogen across agricultural production stages. (Evidence: L1773047508)

  \item Safety, infrastructure readiness, and cost are repeatedly highlighted as non-trivial barriers that
    could limit near-term uptake, even where technical feasibility exists. (Evidence: L1773047103)
\end{itemize}

\subsection*{2.\ Technically Feasible Hydrogen Transition Pathways to 2030}

\begin{description}[style=nextline]
  \item[Pathway A: On-site green hydrogen generation at processing/packing nodes, with leasing/minimal
    upfront capex]
    Rationale: Direct control over hydrogen supply, reduced transport/logistics complexities, and alignment
    with near-term plant-level decarbonization needs. Evidence: L1773047211; L1773047444.
    Constraints: sufficiency of local demand, electricity price conditions, safety/regulatory
    considerations as highlighted in L1773047103.

  \item[Pathway B: Decentralized, small-scale green hydrogen production with local delivery and storage]
    Rationale: Addresses scale-up challenges by spreading capacity across multiple farm/processing sites,
    reducing single-point risk, and enabling modular growth toward 50\% diesel reductions. Evidence:
    L1773047316. Constraints: logistics of distribution, storage losses, and end-use compatibility;
    lifecycle and cost implications as discussed in L1773047508.

  \item[Pathway C: Hydrogen-powered logistics and transport within the apple value chain
    (farm-to-packhouse-to-distribution)]
    Rationale: Direct diesel substitution in mobile and transport activities can yield immediate diesel-use
    reductions if hydrogen-powered fleets or fuel-cell vehicles are adopted. Evidence: L1773047444;
    L1773047211. Constraints: technology readiness for farm/mobile equipment, cost and refueling
    infrastructure, safety/regulatory considerations [L1773047103, L1773047444].

  \item[Pathway D: Integrated hydrogen economy around agri-food hubs with cross-border pilots and
    replication]
    Rationale: Hydrogen valleys and cross-regional pilots can de-risk scale-up by proving integrated
    systems and then replicating across regions. Evidence: L1773047635; L1773047753. Constraints:
    regulatory alignment, funding cycles, and achieving farm-scale readiness to replicate in 2030.

  \item[Pathway E: Policy- and market-accelerated deployment via funding instruments and standards]
    Rationale: A mix of Horizon projects and innovation funds (H2IF, HORIZON demonstrations, regulatory
    services like H2SHIFT) can accelerate deployment patterns adaptable to agri-food systems, potentially
    enabling the diesel-to-H2 transition in apples by 2030. Evidence: L1773047753; L1773047635.
\end{description}

\noindent\textbf{Note on feasibility variability:} The literature consistently stresses that the
transition's success is context-dependent. Factors include hydrogen production/transport/storage
efficiency, end-use technology, electricity mix, and end-user demand economics (L1773047444;
L1773047508).

\subsection*{3.\ Where Perspectives Disagree or Express Uncertainty}

\begin{itemize}
  \item \textbf{Credibility and branding vs.\ technical/economic realism:} The public perspective
    highlights green branding risks and potential greenwashing in hydrogen narratives for agriculture
    (L1773046969; L1773047103).

  \item \textbf{Agri-specific lifecycle data gaps:} Academic entries acknowledge that quantifying lifecycle
    emissions and system efficiency for integrating green hydrogen across multiple agricultural stages is
    not yet well-supported by agri-specific data (L1773047508; L1773047316).

  \item \textbf{Economic viability and deployment timing:} A dairy/steam-generation example (L1773047508)
    indicates that even when decarbonization is technically possible, net present value can be negative
    under certain scenarios, contrasting with entries describing viable commercial models (L1773047211;
    L1773047635; L1773047753).

  \item \textbf{Safety, infrastructure, and regulatory barriers as material blockers:} Several entries
    emphasize safety, storage/transport risks, and regulatory hurdles as critical barriers (L1773047103),
    while others point to on-site/off-site deployment models and pilot programs that could mitigate these
    barriers (L1773047211; L1773047753).
\end{itemize}

\subsection*{4.\ Implications for Decision-Making and Recommended Next Steps}

\begin{itemize}
  \item \textbf{Prioritize near-term pilots} that align with the most feasible, lower-risk deployment
    patterns. Favor on-site generation or decentralized small-scale production with local delivery/storage
    (L1773047211; L1773047316; L1773047103). Tie pilot choices to observed project patterns in EUfunding
    entries (L1773047635).

  \item \textbf{Target heavy-duty transport/hub logistics as a leading edge} for diesel reductions.
    The ledger shows strong focus on hydrogen-enabled heavy-duty transport deployments and related
    refueling infrastructure (L1773047635; L1773047753), which can drive tangible diesel reductions toward
    the 50\% target (L1773047444).

  \item \textbf{Build a controlled, evidence-informed LCA monitoring plan} for agri-specific deployment.
    Given the acknowledged evidence gaps (L1773047508), implement an LCA monitoring framework in pilots to
    quantify actual lifecycle emissions and system efficiency.

  \item \textbf{Leverage EU funding and regulatory pathways} to reduce barriers. Engage with Horizon and
    Innovation Fund structures to access pilots, test-beds, and cross-regional replication patterns
    (L1773047753). Consider linking CyLH2Valley/TH2ICINO-type planning tools and HYPOP public acceptance
    activities.

  \item \textbf{Manage expectations for agri-specific LCA outcomes.} Be prepared for context-dependent
    results: emissions benefits may hinge on a clean electricity mix, efficient hydrogen logistics, and
    farm-level process needs. Frame ``pathway success'' as contingent on supply-chain maturation, policy
    support, and measurable metrics (L1773047508; L1773047444).
\end{itemize}

\noindent\textbf{Key cross-references to consult during planning:}
\begin{itemize}[nosep]
  \item Feasibility and 50\% diesel-reduction framing: L1773047444; L1773047211; L1773047316
  \item Safety, infrastructure, and cost concerns: L1773047103
  \item Agri-LCA data gaps and context-dependency: L1773047508
  \item Commercial models and on-site generation options: L1773047211
  \item Funded pilots and replication opportunities: L1773047635; L1773047753
  \item Regulatory and funding instruments to accelerate deployment: L1773047753
\end{itemize}

\section{Limitations}

The limitations of this approach mainly relate to potential scalability through private data integration
into the vector databases. In such cases, the exposure of sensitive information to external LLM servers
via APIs must be carefully considered with respect to privacy, confidentiality, and data ownership. This
limitation can be mitigated through in-house deployment of the pipeline using locally hosted models;
however, such configurations significantly increase computational requirements for inference compared to
cloud-based deployment~\cite{Li2024}.

Further, with respect to the current datasets, the proposed pipeline offers an evidence-grounded retrieval
mechanism that can remain relatively stable across deployments, but the non-deterministic nature of LLM
inference means that identical prompts may not always produce exactly the same responses.

An additional methodological limitation concerns the retrieval quality diagnostics (e.g., relevance,
coverage, evidence density, and confidence), which are generated by the LLM itself as structured
interpretative signals rather than independently computed validation metrics. These indicators are
intended to enhance transparency regarding the model's perceived grounding strength and evidence use,
but they should not be interpreted as objective performance benchmarks, as they remain sensitive to model
stochasticity, prompt phrasing, and contextual interpretation. Future work could replace or complement
these qualitative diagnostics with deterministic retrieval evaluation metrics and benchmark-based
validation protocols.

In comparison to highly specific technical queries commonly addressed by conventional RAG systems, this
framework extends retrieval-augmented reasoning into broader and more contested domains of socio-technical
knowledge acquisition. As such, increasing domain complexity also increases the need for careful expert
oversight when interpreting outputs. For example, if the LLM is tasked with analysing highly specific
methodological details from technical studies or proprietary engineering documentation, substantially
closer validation of the model's reasoning and interpretation would be required before translating such
outputs into actionable decisions.

Overall, this work aims to establish the foundations of an implementation-oriented post-interpretation
analytical extension to LCA, combining heterogeneous evidence sources as complementary perspectives within
AI-assisted reasoning. Despite the increased complexity, this approach has the potential to significantly
enhance the strategic value of LCA interpretation through scalable integration of technical, industrial,
institutional, and societal knowledge.
